\documentclass[10pt,twocolumn,letterpaper]{article}

\usepackage[pagenumbers]{iccv} 

\definecolor{iccvblue}{rgb}{0.21,0.49,0.74}
\usepackage[pagebackref,breaklinks,colorlinks,allcolors=iccvblue]{hyperref}

\usepackage[accsupp]{axessibility} 

\usepackage{float}
\usepackage{multirow}
\usepackage{multicol}
\usepackage{booktabs}
\usepackage{algorithmic}
\usepackage{algorithm}
\usepackage{array}

\def\paperID{1459} 
\def\confName{ICCV}
\def\confYear{2025}

\title{Alleviating Textual Reliance in Medical Language-guided Segmentation
\\ via Prototype-driven Semantic Approximation}

\author{
Shuchang Ye$^{1}$ \quad
Usman Naseem$^{2}$ \quad
Mingyuan Meng$^{1}$ \quad
Jinman Kim$^{1}$ \\
$^{1}$The University of Sydney, Sydney, Australia \\
$^{2}$Macquarie University, Sydney, Australia \\
{\tt\small shuchang.ye@sydney.edu.au \qquad usman.naseem@mq.edu.au} \\
{\tt\small mmen2292@uni.sydney.edu.au \qquad jinman.kim@sydney.edu.au}
}

\begin{document}

\maketitle

\begin{abstract}
  Medical language-guided segmentation, integrating textual clinical reports as auxiliary guidance to enhance image segmentation, has demonstrated significant improvements over unimodal approaches. However, its inherent reliance on paired image-text input, which we refer to as ``textual reliance", presents two fundamental limitations: 1) many medical segmentation datasets lack paired reports, leaving a substantial portion of image-only data underutilized for training; and 2) inference is limited to retrospective analysis of cases with paired reports, limiting its applicability in most clinical scenarios where segmentation typically precedes reporting. To address these limitations, we propose \textbf{ProLearn}, the first \textbf{Pro}totype-driven \textbf{Learn}ing framework for language-guided segmentation that fundamentally alleviates textual reliance. At its core, in ProLearn, we introduce a novel \textbf{P}rototype-driven \textbf{S}emantic \textbf{A}pproximation (\textbf{PSA}) module to enable approximation of semantic guidance from textual input. PSA initializes a discrete and compact prototype space by distilling segmentation-relevant semantics from textual reports. Once initialized, it supports a query-and-respond mechanism which approximates semantic guidance for images without textual input, thereby alleviating textual reliance. Extensive experiments on QaTa-COV19, MosMedData+ and Kvasir-SEG demonstrate that ProLearn outperforms state-of-the-art language-guided methods when limited text is available. The code is available at \href{https://github.com/ShuchangYe-bib/ProLearn}{https://github.com/ShuchangYe-bib/ProLearn}.
\end{abstract}

\section{Introduction}
\label{sec:introduction}

\begin{figure}[t]
  \centering
   \includegraphics[width=\linewidth]{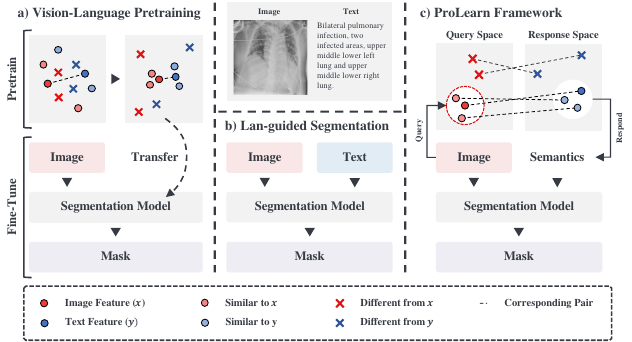}
   \caption{Comparison of vision-language paradigms for medical image segmentation. a) VLP: Pretrained on large-scale image-text pairs and then fine-tuned on image-only data from the target dataset. b) Language-guided Segmentation: Requires strict image-text pairs during both training and inference. c) ProLearn: Initializes with a limited amount of paired image-text data to construct query-response spaces. After initialization, it enables learning with limited textual input and performs inference without text.}
\label{fig:diagram_comparison}
\end{figure}

Segmentation is an essential tool in medical image analysis, supporting clinical workflows by enabling precise delineation of anatomical structures, identification of pathological regions, and facilitating targeted interventions~\cite{medsam}. Its applications span critical tasks such as disease diagnosis, treatment planning, and surgery support~\cite{unet_review24}. Deep learning has revolutionized segmentation, making it more accurate, reliable, and widely applicable in clinical practice. Unimodal (image-only) segmentation methods~\cite{unet_review22}, such as U-Net~\cite{unet} and its extensions~\cite{unet_review21}, including U-Net++~\cite{unet++}, Attention U-Net~\cite{att_unet}, and Trans U-Net~\cite{trans_unet}, have been widely adopted in medical imaging. In recent years, multimodal segmentation methods that leverage textual clinical reports as complementary information have gained wide attention for their ability to transcend the performance limits of unimodal segmentation~\cite{vlm_survey24, vlm_survey23, vlm_sruvey22}.

The exploration of multimodal segmentation~\cite{vlm_survey24, vlm_survey23, vlm_sruvey22} began with vision-language pretraining (VLP)~\cite{clip}, as shown in Figure~\ref{fig:diagram_comparison}a, where models pre-trained on paired textual and visual data demonstrated improved visual understanding and performance when finetuned on target image-only segmentation datasets~\cite{clip_prompt, clip_ref, clip_seg_tune}. However, these pretraining-based approaches fail to fully exploit the disease-specific information embedded in the target dataset’s reports, as the general knowledge learned during pretraining often lacks the domain-specific details essential for precise disease lesion segmentation. To address this limitation, language-guided segmentation has been proposed as a promising approach~\cite{lvit, guidedecoder, madapter, lga}, which takes textual clinical report inputs as auxiliary semantic guidance for image segmentation (see Figure~\ref{fig:diagram_comparison}b). It has achieved remarkable success by leveraging textual abnormality description as explicit semantic guidance to segmentation, thus outperforming pretraining-based approaches and setting new benchmarks in medical image segmentation.

However, language-guided segmentation methods have inherent reliance on paired image-text data due to the requirement for textual reports as auxiliary inputs. This reliance is referred to as textual reliance in this study and has incurred notable limitations during the training and inference stages. In the training stage, its reliance on paired image-text data leads to the underutilization of image-only data, which constitutes the majority of available datasets~\cite{universeg}. In the inference stage, reliance on textual descriptions during inference confines its use to retrospective analysis, misaligning with most clinical workflows where segmentation is required preemptively for tasks such as preoperative planning~\cite{seg_preoperative_planning}, diagnostic decision making~\cite{seg_brain, seg_breast, seg_lung}, and real-time procedural guidance~\cite{seg_therapy, seg_vr, seg_needle}.

\begin{figure}[t]
  \centering
   \includegraphics[width=\linewidth]{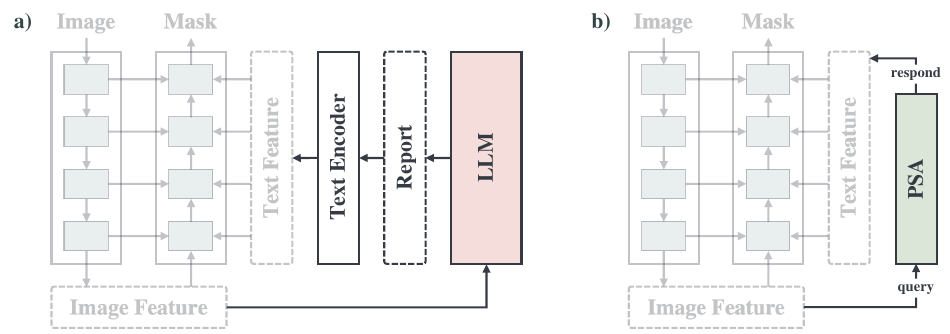}
   \caption{Flow diagram comparison between SGSeg~\cite{sgseg} and our proposed ProLearn. a) SGSeg: \textit{image $\rightarrow$ LLM $\rightarrow$ report $\rightarrow$ BERT $\rightarrow$ embedding}. b) ProLearn: \textit{image $\rightarrow$ PSA $\rightarrow$ embedding}.}
\label{fig:sgseg_v_prolearn}
\end{figure}

Our previous study, Self-Guided Segmentation (SGSeg)~\cite{sgseg}, made a preliminary attempt to eliminate textural reliance during inference in language-guided segmentation by leveraging large language models (LLMs) to generate synthetic reports to compensate for the missing textual input, as illustrated in Figure~\ref{fig:sgseg_v_prolearn}a. However, the integration of LLMs substantially increases the model size and inference time, making the approach unsuitable for deployment on edge devices~\cite{edge} or for real-time applications such as image-guided surgery~\cite{seg_sur}. Moreover, textual reliance during training remains an open issue.

We argue that the critical guidance in language-guided segmentation is not the entire clinical report, which is often verbose and cluttered with irrelevant information, but rather the specific segmentation-relevant semantic features embedded within it. Our investigation further indicates that the semantic space of medical reports is inherently constrained, consisting of a finite set of distinct representations relevant to segmentation tasks, as clinical reports typically adhere to standardized medical terminologies, which result in a relatively closed vocabulary.

Building on these insights, we propose \textbf{ProLearn} (see Figure~\ref{fig:diagram_comparison}c), a lightweight and efficient \textbf{Pro}totype \textbf{Learn}ing framework that fundamentally alleviates textual reliance during both training and inference. ProLearn introduces a \textbf{P}rototype-Driven \textbf{S}emantic \textbf{A}pproximation (\textbf{PSA}) module, which enables the model to approximate semantic guidance without the need for textual input. Specifically, PSA constructs a discrete and compact prototype space by distilling segmentation-relevant concepts from clinical reports. It then provides a query-and-respond mechanism to support interaction between segmentation models and the prototype space, where unseen semantics are approximated by weighted aggregation of the existing prototypes based on similarity. Therefore, PSA enables segmentation models to query by image feature and receive responded semantics feature as guidance for feature maps refinement, as illustrated in Figure~\ref{fig:sgseg_v_prolearn}b. 

\noindent The main contributions of this work are as follows:
\begin{itemize}
    \item To the best of our knowledge, our proposed ProLearn is the first work to alleviate textual reliance in medical language-guided segmentation in both training and inference.
    \item We introduce a novel PSA module that supports learning with both paired image-text data and image-only data while enabling inference without textual input by querying a learned prototype space to provide semantic guidance for segmentation.
    \item Extensive experiments on QaTa-COV19, MosMedData+ and Kvasir-SEG demonstrate that ProLearn outperforms language-guided segmentation methods in limited text availability settings (see Section~\ref{sec:sota_comparison}) and surpasses state-of-the-art unimodal segmentation models (see Section~\ref{sec:tf_comparison}). Compared to SGSeg, ProLearn achieves a $1000\times$ reduction in parameters and $100\times$ faster inference speed (see Section~\ref{sec:complexity_comparison}).
\end{itemize}

\begin{figure*}[t]
\begin{center}
\includegraphics[width=0.85\linewidth]{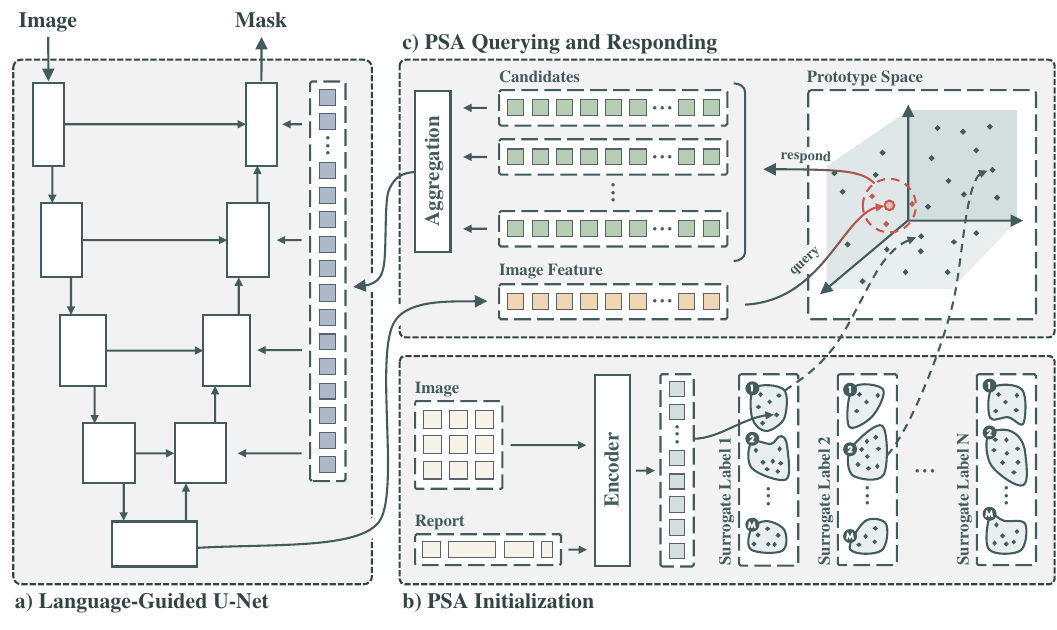}
\end{center}
   \caption{The overview of the ProLearn framework for alleviating textual reliance in medical language-guided segmentation. a) Language-guided U-Net: The U-Net encoder extracts image features, which are used to query the PSA module. The decoder is then guided by PSA's response to decode the segmentation mask. b) PSA Initialization: Clinical reports are processed and abstracted into a discrete prototype space, representing segmentation-relevant semantic and spatial information. c) PSA Querying and Responding: PSA receives image feature queries, selects relevant prototype candidates, and responds with an aggregated representation to approximate the pathological region's segmentation-relevant semantic representation.}
\label{fig:prolearn}
\end{figure*}


\section{Related Work}
\label{sec:related_work}

\subsection{Language-guided Segmentation}

Language-guided segmentation aims to address the gap that the target dataset's reports are not fully exploited for learning in conventional unimodal~\cite{unet, unet++, att_unet, trans_unet, swin_unet} or VLP methods~\cite{clip, clip_prompt, clip_ref, clip_seg_tune}. LViT~\cite{lvit} was the first work in language-guided segmentation, which takes both images and textual reports as input to train a multimodal-input segmentation model. They annotated existing publicly available segmentation datasets, QaTa-COV19~\cite{qata} and MosMedData+~\cite{mosmeddata}, with corresponding clinical reports. To fuse the text features with feature maps in the U-Net, LViT adopted an early fusion approach, which introduced a Pixel-Level Attention Module (PLAM) to involve textual features as semantic guidance. LViT showed consistent performance over image-only and VLP segmentation methods, which demonstrated the importance and potential of textual semantic guidance from clinical reports in target datasets.

Subsequent methods proposed more flexible and robust feature fusion modules for language-guided segmentation. GuideSeg~\cite{guidedecoder} moved away from early fusion while adopting a late-fusion strategy that fused textual and visual features at the decoding stage, where the text features were better preserved and more effectively influenced the segmentation process. MAdaptor~\cite{madapter}  addressed the unidirectional flow of textual semantics seen in previous frameworks. It introduced a bidirectional adaptor connecting multiple layers of unimodal encoders, facilitating mutual information exchange between text and image representations at various scales. LGA~\cite{lga} adopted a parameter-efficient fine-tuning strategy that preserved the original parameters of large segmentation foundation models~\cite{medsam}. These works further prove that integrating target datasets' textual reports as guidance can significantly improve the performance of segmentation models. 

These methods suffered from textual reliance. The reliance during training restricted the utilization of large portions of medical datasets that lacked paired textual reports. The reliance during inference limits their practical utility in most clinical scenarios using segmentation without reports. SGSeg~\cite{sgseg} attempted to release the textual reliance at inference by training a LLM~\cite{llm_med} to generate needed clinical reports from images. Nevertheless, the inclusion of LLMs increased model size and inference time, making it unsuitable for edge devices and real-time applications. The textual reliance during training still remained unsolved. Our proposed ProLearn fundamentally alleviates textual reliance in training and inference, and its prototype design PSA significantly reduces the number of parameters and inference time compared to LLM-based SGSeg.

\subsection{Prototype Learning}

Prototype learning draws on the principle that images and text can be effectively captured in discrete prototype representations rather than relying on freshly built embeddings for every instance~\cite{discrete_rl}. Each prototype functions as a canonical reference, reflecting the typical features of its corresponding class. Such a strategy is widely adopted in classification~\cite{prototype_class}, where an unseen query is associated with the class whose prototype lies closest in a learned space. By prioritizing the reuse of well-established representations, prototype-based methods frequently attain top-tier performance under data-scarce settings~\cite{prototype_fewshot}. Beyond unimodal applications, prototype learning has demonstrated its potential in multimodal tasks, effectively aligning features across modalities~\cite{prior}, such as images and text. Inspired by the principles of expressing unseen queries via finite existing prototypes, the proposed ProLearn extracts semantic information from textual reports as prototypes, enabling representing semantic guidance from existing image-text pairs for image-only input during training and inference, with improved parameter, computation, and data efficiency.

\section{Methodology}
\label{sec:method}

Figure~\ref{fig:prolearn} shows the overall architecture of ProLearn, a framework that leverages an efficient prototype learning module called \textbf{P}rototype-\textbf{D}riven \textbf{S}emantic \textbf{A}pproximation (\textbf{PSA}) alongside a Language-guided U-Net backbone for segmentation. PSA is designed to (i) reduce the need for textual annotations during training by using available text only once to initialize a discrete prototype space and (ii) remove the need for textual input entirely at inference. 

\subsection{PSA Initialization}
The PSA initialization is a one-time process that constructs a queryable prototype space before training. Given that a training set includes $K$ paired image-text samples and other image-only samples, where the $i$-th sample is denoted as $\langle I_i, T_i \rangle$, with $I_i$ as the image and $T_i$ as its associated report. Each paired sample is processed by a trained domain-specific vision-language encoder (BioMedCLIP~\cite{biomedclip}), $f_{\text{enc}}^I$ and $f_{\text{enc}}^T$ to extract their semantic features $e^I_i$ and $e^T_i$:
\begin{equation}
    e^I_i = f_{\text{enc}}^I(I_i), \quad e^T_i = f_{\text{enc}}^T(T_i), \quad e^I_i,e^T_i \in \mathbb{R}^D
\end{equation}

\begin{figure}[t]
  \centering
   \includegraphics[width=\linewidth]{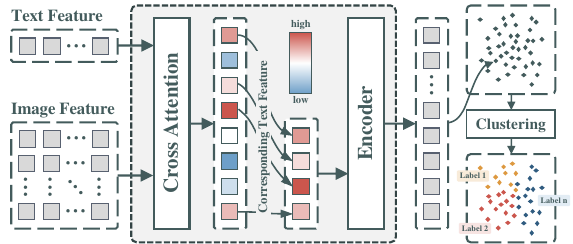}
   \caption{Attention-guided surrogate label extraction.}
\label{fig:surrogate_label}
\end{figure}

\textbf{Surrogate Label Extraction}:
To extract finite segmentation-relevant semantics in the clinical reports, we cluster the image-text pairs into $N$ semantic surrogate labels, as shown in Figure~\ref{fig:surrogate_label}.

As many clinical reports are verbose, we first isolate the tokens most relevant to segmentation. To achieve this, we utilize the cross-attention module of a separately trained Language-guided U-Net, which has the same architecture as the language-guided U-Net in Figure~\ref{fig:prolearn}, to assess token relevance in all $K$ available image-text pairs $\{\langle I_i, T_i \rangle \mid i=1, \dots, K\}$. Specifically, for each text input $T_i$ paired with an image $I_i$, the cross-attention weight $\alpha_j$ computed for each token $t_j$ in $T_i$ indicates the relevance of each token $t_j$ in guiding the segmentation of $I_i$. Tokens whose attention scores exceed a threshold $\tau$ are retained, resulting in a shorter segmentation-relevant sentence for each $T_i$, which we denote $T^{\text{selected}}_i$:
\begin{equation}
    T^{\text{selected}}_i = \{t_j \mid \alpha_{j} > \tau, t_j\in T_i\}.
\end{equation}
We then feed $T^{\text{selected}}_i$ into the text encoder $f_{\text{enc}}^T$, obtaining a semantic feature $e^\text{sem}_i$:
\begin{equation}
    e^\text{sem}_i = f_{\text{enc}}^T\left(T^{\text{selected}}_i\right), \quad \forall i \in \{1, \dots, K\}
\end{equation}

We then group similar textual semantics together using the hierarchical density-based clustering algorithm, HDBSCAN~\cite{hdbscan}. The clustering results in $N$ surrogate labels $\{l_1, l_2, ..., l_N\}$. Each surrogate label $l_i$ encapsulates a distinct segmentation-relevant textual semantics, with its corresponding cluster $\mathcal{C}_i$ representing the grouped features:
\begin{equation}
	\mathcal{C}_i = \{ \langle e^I_j, e^T_j \rangle \mid \operatorname{HDBSCAN}(e^\text{sem}_i) = l_i \}.
\end{equation}

\textbf{Prototype Space Construction}:
While the above surrogate labels capture sparse textual semantics, medical images often convey richer and more fine-grained information. To make the prototype space compact and more representative for image queries, each text-based surrogate label cluster $\mathcal{C}_i$ is further subdivided into sub-clusters $\mathcal{C}_{ij}$ via K-means clustering algorithm~\cite{kmeans}:
\begin{equation}
    \mathcal{C}_{ij} = \{ \langle e^I_k, e^T_k \rangle \mid \operatorname{K-Mean}(e^I_k) = j, \langle e^I_j, e^T_j \rangle \in \mathcal{C}_i\}.
\end{equation}
For each sub-cluster $\mathcal{C}_{ij}$, instead of taking the centroid itself, we locate the sample $c_{ij}$ closest to the cluster's centroid and treat it as a fine representation of disease lesion. This approach reduces the influence of outliers that can significantly distort the centroid. Given $c_{ij} = \langle e^I_k, e^T_k \rangle$, we assign its containing image feature $e^I_k$ as visual query prototype $q_{ij}$ and text feautre $e^T_k$ as textual respond prototype $r_{ij}$. By repeating this across all sub-clusters, we construct a discrete and compact initial prototype space $\mathcal{S}$, consisting of a query space $\mathcal{S}^Q$ and a response space $\mathcal{S}^R$, both of dimension $N \times M \times D$: 
\begin{equation}
    \mathcal{S} = \left( \mathcal{S}^Q, \mathcal{S}^R \right) = \bigcup_{i=1}^{N} \bigcup_{j=1}^{M} \langle q_{ij}, r_{ij} \rangle.
\end{equation}
Each visual query prototype \(q_{ij}\) in \(\mathcal{S}^Q\) is directly linked to its corresponding textual response prototype \(r_{ij}\) in \(\mathcal{S}^R\):
\begin{equation}
    q_{ij} \longrightarrow r_{ij}, \quad \forall i \in \{1, \dots, N\}, \quad j \in \{1, \dots, M\}.
\end{equation}
This prototype space is dynamically learned during the training process of ProLearn.

\subsection{PSA Querying and Responding}

During both training and inference, the Language-guided U-Net $f_{\text{seg}}$ queries the query space $\mathcal{S}^Q$ using an image feature $q^*$, encoded by the image encoder $f^I_{\text{enc}}$ from an image input $I^*$, enabling the use of image-only samples. Through the query-and-response mechanism of the PSA module, it responds with an approximate textual semantic feature $r^*$ which guides $f_{\text{seg}}$ without textual input.

\textbf{PSA Querying}: The PSA querying process involves searching the most relevant visual queries from the query prototype space $\mathcal{S}^Q$. Given the encoded image feature $q*$, the PSA module computes the cosine similarity scores $s_{ij}$ between $q^*$ and each query prototype $q_{ij}$:
\begin{equation}
    s_ij = s(q^*, q_{ij}) = \frac{q^* \cdot q_{ij}}{\|q^*\| \|q_{ij}\|}
\end{equation}
After ranking these similarity scores, the PSA module selects the top-$k$ query prototypes $Q^*$ that best match $q^*$:
\begin{equation}
    Q^* = \{q_{ij} \mid \forall i,j \in \operatorname{arg}\operatorname{top}_k (s_{ij})\}.
\end{equation}

\textbf{PSA Responding}: The PSA responding process finds the corresponding response prototypes $R^*$ that are linked to the selected query prototypes $Q^*$:
\begin{equation}
    R^* = \{r_{ij} \mid q_{ij} \in Q^*, q_{ij} \longrightarrow r_{ij}\}.
\end{equation}
These response prototypes $R^*$ are referred to as candidates in Figure~\ref{fig:prolearn}b. The candidates are then aggregated using a similarity-weighted sum, where the weights are computed through the softmax function over the similarity scores:
\begin{equation}
    r^* = \sum_{r_i \in R^*} w_i r_i, \quad w_i = \frac{\exp(s(q^*, q_i))}{\sum_{q_j \in Q^*} \exp(s(q^*, q_j))}.
\end{equation}
At this point, the PSA responding process is complete. We then feed $r^*$ into the decoding process of Language-guided U-Net, providing an approximated semantic guidance.

\section{Experimental Setup}
\label{sec:experiments}

\begin{table*}
\begin{center}
\scriptsize
\renewcommand{\arraystretch}{1.2}
\begin{tabular}{llccccc|ccccc}
    \toprule
    \textbf{Dataset} & \textbf{Model} & \multicolumn{5}{c|}{\textbf{Dice}} & \multicolumn{5}{c}{\textbf{mIoU}} \\
    \cmidrule(lr){3-7} \cmidrule(lr){8-12}
    & & 50\% & 25\% & 10\% & 5\% & 1\% & 50\% & 25\% & 10\% & 5\% & 1\% \\
    \midrule
    \multirow{4}{*}{QaTa-COV19~\cite{qata}} 
    & LViT~\cite{lvit} & 0.8416 & 0.8202 & 0.8004 & 0.7638 & 0.7006 & 0.7320 & 0.7012 & 0.6747 & 0.6248 & 0.5490 \\
    & GuideSeg~\cite{guidedecoder} & 0.8633 & 0.8524 & 0.8402 & 0.8240 & 0.7333 & 0.7595 & 0.7428 & 0.7244 & 0.7007 & 0.5789 \\
    & SGSeg~\cite{sgseg} & 0.8641 & 0.8574 & 0.8423 & 0.8057 & 0.7307 & 0.7607 & 0.7504 & 0.7276 & 0.6746 & 0.5757 \\
    & \textbf{ProLearn} & \textbf{0.8667} & \textbf{0.8598} & \textbf{0.8583} & \textbf{0.8573} & \textbf{0.8566} & \textbf{0.7721} & \textbf{0.7690} & \textbf{0.7573} & \textbf{0.7558} & \textbf{0.7553} \\
    \midrule
    \multirow{4}{*}{MosMedData+~\cite{mosmeddata}} 
    & LViT~\cite{lvit} & 0.7189 & 0.6608 & 0.5501 & 0.5069 & 0.1677 & 0.5696 & 0.5042 & 0.4108 & 0.3538 & 0.1015 \\
    & GuideSeg~\cite{guidedecoder} & 0.7508 & 0.7393 & 0.6898 & 0.6375 & 0.4235 & 0.6089 & 0.5864 & 0.5265 & 0.4678 & 0.2686 \\
    & SGSeg~\cite{sgseg} & 0.7455 & 0.7439 & 0.6950 & 0.6465 & 0.3452 & 0.5943 & 0.5922 & 0.5325 & 0.4776 & 0.2086 \\
    & \textbf{ProLearn} & \textbf{0.7539} & \textbf{0.7512} & \textbf{0.7424} & \textbf{0.7379} & \textbf{0.7218} & \textbf{0.6126} & \textbf{0.6109} & \textbf{0.6087} & \textbf{0.6069} & \textbf{0.6032} \\
    \midrule
    \multirow{4}{*}{Kvasir-SEG~\cite{kvasir}} 
    & LViT~\cite{lvit}          
    & 0.7669 & 0.6424 & 0.5719 & 0.5482 & 0.4272 
    & 0.6228 & 0.4769 & 0.4025 & 0.3782 & 0.2729 \\
    & GuideSeg~\cite{guidedecoder}      
    & 0.8848 & 0.8390 & 0.7754 & 0.7497 & 0.5615 
    & 0.7939 & 0.7228 & 0.6360 & 0.6043 & 0.4008 \\
    & SGSeg~\cite{sgseg}         
    & 0.8769 & 0.8304 & 0.8025 & 0.7526 & 0.5406 
    & 0.7808 & 0.7099 & 0.6702 & 0.6034 & 0.3705 \\
    & \textbf{ProLearn}    
    & \textbf{0.8983} & \textbf{0.8946} & \textbf{0.8898} & \textbf{0.8823} & \textbf{0.8718} 
    & \textbf{0.8162} & \textbf{0.8101} & \textbf{0.8020} & \textbf{0.7905} & \textbf{0.7729} \\
    \bottomrule
\end{tabular}
\end{center}
\caption{Performance comparison of language-guided segmentation models in simulated scenarios with limited text supervision on the QaTa-COV19, MosMedData+ and Kvasir-SEG dataset. The
best results are highlighted in bold.}
\label{tab:sota_comparison}
\end{table*}

\subsection{Datasets}

To evaluate our approach, we used the two well-benchmarked datasets: QaTa-COV19~\cite{qata}, MosMedData+~\cite{mosmeddata} and Kvasir-SEG~\cite{kvasir}, which are commonly adopted for performance comparison in language-guided segmentation~\cite{lvit, guidedecoder, sgseg, lga, madapter}.
 
\textbf{QaTa-COV19} is a large-scale dataset consisting of $9{,}258$ chest X-ray images with manually annotated COVID-19~\cite{covid} lesion masks. To facilitate language-guided segmentation, LViT~\cite{lvit} extends this dataset with textual descriptions detailing bilateral lung infections, the number of affected regions, and their spatial localization within the lungs. We adopt the official dataset split: $5{,}716$ images for training, $1{,}429$ for validation, and $2{,}113$ for testing.

\textbf{MosMedData+} comprises $2{,}729$ CT slices depicting pulmonary infections. Similar to QaTa-COV19, LViT augments this dataset with text-based annotations to support language-guided segmentation tasks. We adopt the official dataset split: $2{,}183$ slices for training, $273$ for validation, and $273$ for testing.

\textbf{Kvasir-SEG} is a publicly available dataset comprising $1{,}000$ colonoscopy images of gastrointestinal polyps with corresponding pixel-level segmentation masks. We follow a standard 8:1:1 split for training, validation, and testing.

\subsection{Evaluation Metrics}

To quantitatively evaluate segmentation performance, we used the metrics which are used in the previous similar studies~\cite{lvit, guidedecoder, sgseg, lga, madapter}: Dice coefficient~\cite{dice} and the mean Intersection over Union (mIoU~\cite{miou}), two widely used metrics for measuring spatial overlap between predicted and ground truth segmentation masks. 
The formulations are defined in Equations~\ref{eq:dice} and~\ref{eq:miou}.
\begin{equation}
    \text{Dice} = \frac{1}{N} \sum_{i=1}^{N} \frac{2 \sum_{c=1}^{C} | P_{i}^{(c)} \cap G_{i}^{(c)} |}{\sum_{c=1}^{C} \big( |P_{i}^{(c)}| + |G_{i}^{(c)}| \big) }
    \label{eq:dice}
\end{equation}
\begin{equation}
    \text{mIoU} = \frac{1}{N} \sum_{i=1}^{N} \frac{1}{C} \sum_{c=1}^{C} \frac{| P_{i}^{(c)} \cap G_{i}^{(c)} |}{| P_{i}^{(c)} \cup G_{i}^{(c)} |}
    \label{eq:miou}
\end{equation}
where \( N \) represents the number of images in the dataset, \( C \) denotes the number of semantic categories, and \( P_{i}^{(c)} \) and \( G_{i}^{(c)} \) correspond to the predicted and ground truth segmentation masks for class \( c \) in image \( i \), respectively. 

\subsection{Experimental Design}

To evaluate ProLearn, we compare it against both multimodal and unimodal segmentation methods, analyzing its performance under various realistic clinical scenarios.

\textbf{Limited availability of text}: To illustrate the limitations of language-guided segmentation methods that require strict image-report pairing and highlight the importance of training with both image-text-paired and image-only data, we simulate a real-world clinical scenario where detailed radiology reports are often unavailable for a significant portion of the dataset. Due to the lack of large-scale benchmarks containing both paired and unpaired examples, we construct a counterfactual setting by progressively reducing the proportion of image-text pairs in the training data to $50\%$, $25\%$, $10\%$, $5\%$, and $1\%$, while discarding the remaining unpaired images to simulate their inaccessibility. We compare ProLearn against state-of-the-art language-guided segmentation models: LViT~\cite{lvit}, GuideSeg~\cite{guidedecoder}, and SGSeg~\cite{sgseg}.

\textbf{Segmentation in real-world (image-only) setting}: In the majority of clinical scenarios, such as real-time procedural guidance and decision support, segmentation is used without text. To align with real-world settings, we evaluate ProLearn in a strictly ``image-only" setting, where no text input is provided during either training or inference on the target datasets. Under these conditions, we compare ProLearn to established unimodal segmentation methods and vision-language pretraining models (CLIP~\cite{clip} and GLoRIA~\cite{gloria}) adapted for image-only use.

\textbf{Prototype vs. LLM}: To demonstrate our PSA's advantages against LLMs in real-world deployment, we focus on inference speed, model size, and time complexity. We specifically compare ProLearn with SGSeg~\cite{sgseg}, a recent method that leverages LLMs (e.g., GPT-2~\cite{gpt2}, Llama3~\cite{llama3-med}) to generate synthetic textual reports at inference in order to compensate for missing textual input.

\textbf{Qualitative Analysis}: We provide visual comparisons against state-of-the-art language-guided segmentation models under progressively lower text availability settings. Specifically, We visualize both final segmentation outputs and corresponding saliency maps.

\textbf{Hyperparameter Sensitivity Analysis}: We analyze how model performance varies as we adjust two key hyperparameters: the number of prototypes $M$, which governs the compactness of the prototype space, and the number of candidate responses $k$.

\section{Results and Discussion}
\label{sec:results&discussion}

\subsection{Comparison with State-Of-The-Art Methods}
\label{sec:sota_comparison}

\begin{figure}[t]
  \centering
   \includegraphics[width=0.95\linewidth]{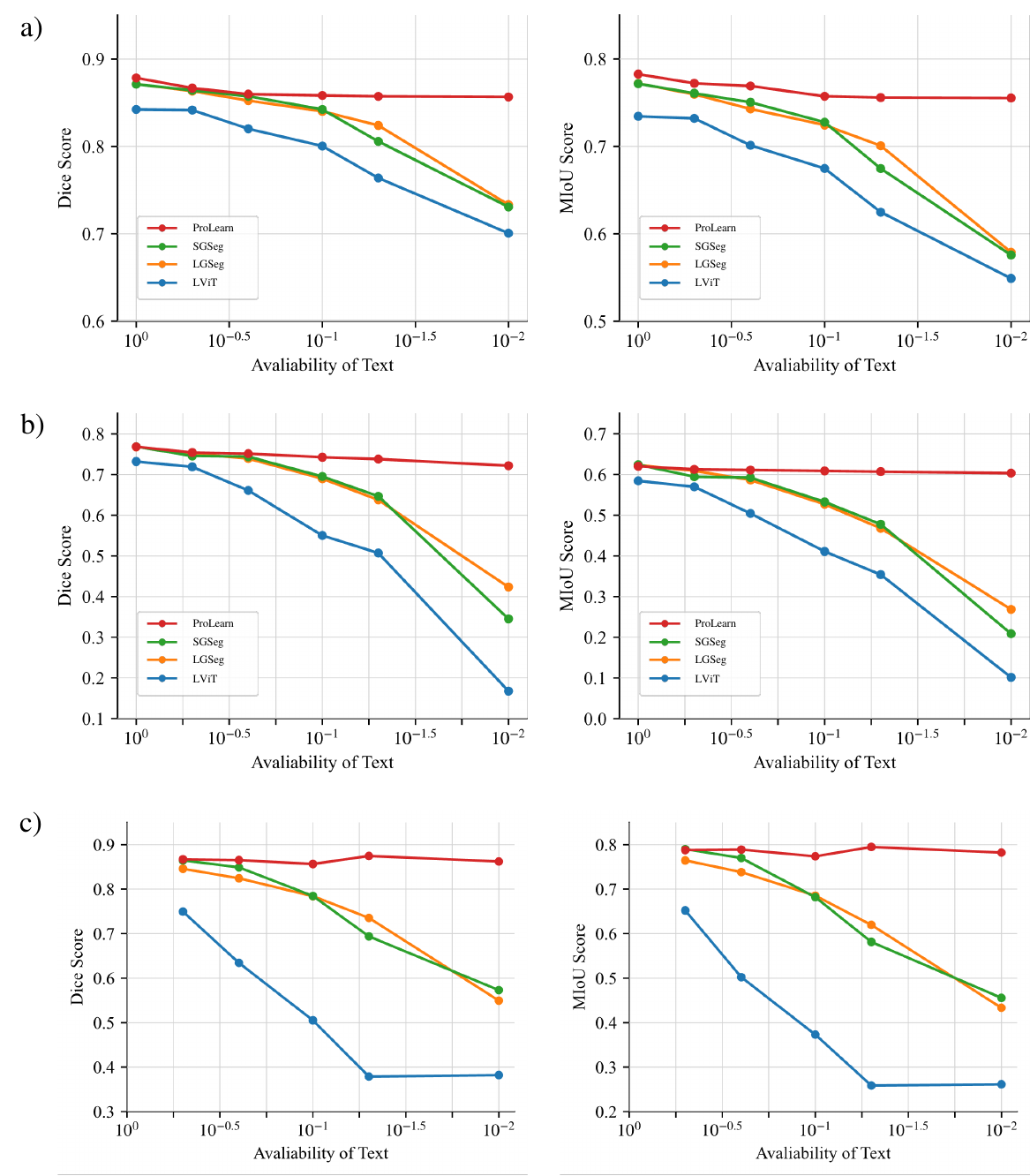}
   \caption{Performance degradation as the ratio of text availability decreases. The plots show Dice and nIoU metrics for a) the QaTa-COV19, b) the MosMedData+ and c) the Kvasir-SEG dataset.}
\label{fig:degradation}
\end{figure}

Table~\ref{tab:sota_comparison} presents our comparative results under clinical scenarios where paired textual reports are limited ($50\%$, $25\%$, $10\%$, $5\%$, and $1\%$). ProLearn achieved higher Dice and mIoU scores than existing language-guided methods in all settings, and this performance gap increased as the fraction of available text decreased. For example, in the $1\%$ text in MosMedData+, ProLearn retained a Dice score of $0.7218$, compared to SGSeg $0.3452$ and LViT $0.1677$. This trend underscores the effectiveness of ProLearn in learning from image-text and image-only data, allowing it to outperform approaches that require textual input for both training and inference. Although SGSeg also requires no text at inference, its performance remained below ProLearn due to its limited capacity to exploit learned semantics when text is scarce. Further discussions on the comparison between SGSeg and ProLearn can be found in Section~\ref{sec:complexity_comparison}.

The performance degradation analysis, visualized in Figure~\ref{fig:degradation}, further illustrates the importance of reducing textual reliance in training. ProLearn exhibited the most minor degradation, preserving robust performance even under severely limited text supervision. 

\subsection{Comparison in real-world (image-only) setting}
\label{sec:tf_comparison}

We further evaluated the effectiveness of ProLearn's use of a target dataset's auxiliary text under near ``image-only" conditions, as most clinical workflows require segmentation models operating without textual input. As shown in Table~\ref{tab:unimodal_comparison}, ProLearn outperformed both unimodal approaches (U-Net, Attention U-Net, Swin U-Net, etc.) and well-recognized VLP methods (CLIP, GLoRIA), even when only $10\%$, $5\%$, or $1\%$ of the paired reports were available. This demonstrates the importance of textual semantics within the target dataset. Unlike unimodal and VLP-based methods, which fail to leverage clinical reports during fine-tuning on target datasets, ProLearn incorporates available image-report pairs in the learning process.

\begin{table}
\scriptsize
\begin{center}
\renewcommand{\arraystretch}{1.2}
\begin{tabular}{lcc|cc|cc}
    \toprule
    \textbf{Model} & \multicolumn{2}{c|}{\textbf{QaTa-COV19}} & \multicolumn{2}{c}{\textbf{MosMedData+}} & \multicolumn{2}{c}{\textbf{Kvasir-SEG}} \\
    \cmidrule(lr){2-3} \cmidrule(lr){4-5} \cmidrule(lr){6-7}
    & \textbf{Dice} & \textbf{mIoU} & \textbf{Dice} & \textbf{mIoU} & \textbf{Dice} & \textbf{mIoU} \\
    \midrule
    U-Net~\cite{unet} & 0.819 & 0.692 & 0.638 & 0.505 & 0.195 & 0.182 \\
    U-Net++~\cite{unet++} & 0.823 & 0.706 & 0.714 & 0.582 & 0.280 & 0.180 \\
    Attention U-Net~\cite{att_unet} & 0.822 & 0.701 & 0.664 & 0.528 & 0.364 & 0.226 \\
    Trans U-Net~\cite{trans_unet} & 0.806 & 0.687 & 0.702 & 0.575 & 0.048 & 0.100 \\
    Swin U-Net~\cite{swin_unet} & 0.836 & 0.724 & 0.669 & 0.531 & 0.398 & 0.246 \\
    CLIP~\cite{clip} & 0.798 & 0.707 & 0.720 & 0.596 & -- & -- \\
    GLoRIA~\cite{gloria} & 0.799 & 0.707 & \textbf{0.722}  & 0.602 & -- & -- \\
    MedSAM~\cite{medsam} & 0.730 & 0.619 & 0.509  & 0.371 & -- & -- \\
    BiomedParse~\cite{biomedparse} & 0.781 & 0.682 & 0.671  & 0.553 & 0.828 & 0.721 \\
    \textbf{ProLearn (1\%)} & \textbf{0.857} & \textbf{0.755} & \textbf{0.722} & \textbf{0.603} & \textbf{0.872} & \textbf{0.773} \\
    \textbf{ProLearn (5\%)} & \textbf{0.857} & \textbf{0.756} & \textbf{0.738} & \textbf{0.607} & \textbf{0.882} & \textbf{0.790} \\
    \textbf{ProLearn (10\%)} & \textbf{0.858} & \textbf{0.757} & \textbf{0.742} & \textbf{0.609} & \textbf{0.889} & \textbf{0.802} \\
    \bottomrule
\end{tabular}
\end{center}
\caption{Performance comparison of methods in \textit{image-only settings} where paired reports are excluded entirely from the target datasets' training and inference. The best results are highlighted in bold.}
\label{tab:unimodal_comparison}
\end{table}

\subsection{Prototype vs. LLM}
\label{sec:complexity_comparison}

\begin{figure}[t]
  \centering
   \includegraphics[width=\linewidth]{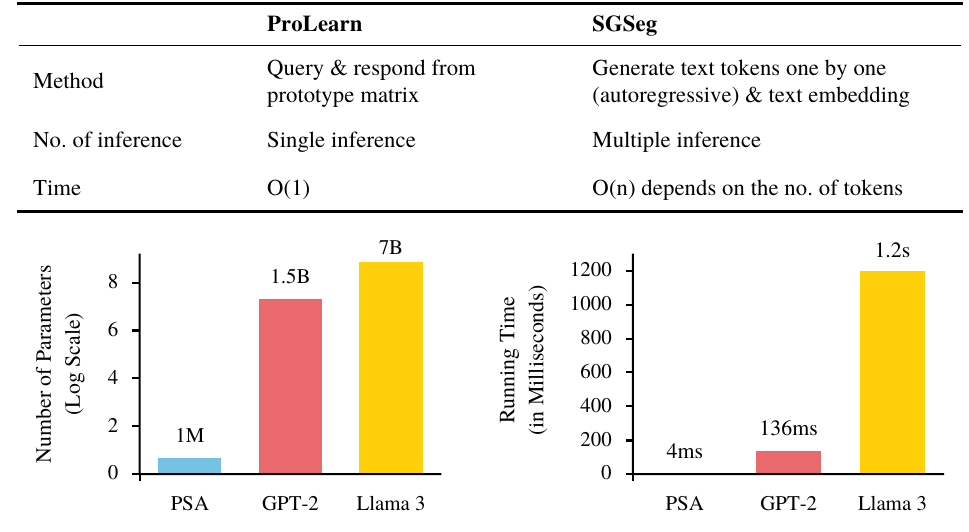}
   \caption{Comparison between the proposed prototype-driven approach, ProLearn, and the LLM-based approach, SGSeg. Top: The theoretical analysis of time complexity. Bottom: The model size and inference time comparison.}
\label{fig:complexity}
\end{figure}

Prototype learning enables constant-time inference of $\mathcal{O}(1)$ because it uses a pre-defined, finite prototype space that is independent of both image and text size. At the same time, LLM applies autoregressive image-to-text generation, which has a linear time complexity $\mathcal{O}(n)$, where $n$ represents the number of tokens in the generated text (see Figure~\ref{fig:complexity}, top). As shown in Figure~\ref{fig:complexity}, bottom, prototype learning achieves an inference time of 4ms, making it $100\times$ faster than GPT-2 (136ms) and $300\times$ faster than Llama3 (1.2s). Moreover, the prototype model has only $1$M parameters, which is $1000\times$ smaller than large language models such as GPT-2 (1.5B parameters) and Llama3 (7B parameters). These properties make prototype learning highly efficient for real-time applications and suitable for deployment on resource-constrained edge devices.

\begin{figure*}[t]
\begin{center}
\includegraphics[width=\linewidth]{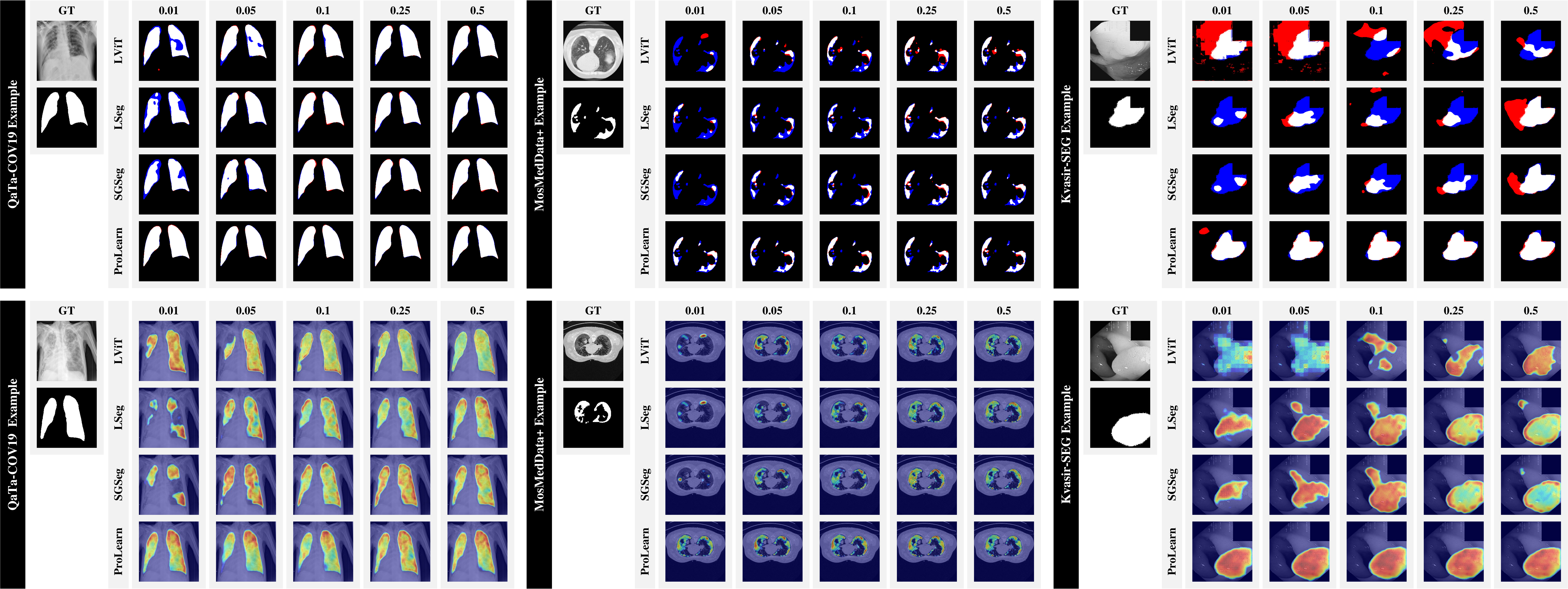}
\end{center}
   \caption{Visualization comparison of language-guided segmentation methods under different text availabilities. The upper row shows segmentation outputs on CXR\_S09687-E18404-R1 from the QaTa-COV19 dataset (left), Jun\_radiopaedia\_4\_85506\_1\_case19\_16 from the MosMedData+ dataset (middle) and ju5xkwzxmf0z0818gk4xabdm from the Kvasir-SEG dataset (right). Blue regions indicate ground-truth pixels missed by the model, while red regions indicate pixels mistakenly predicted as lesion. The lower row presents the saliency map interpretability study for different approaches on CXR\_S09346-E23164-R1 from the QaTa-COV19 dataset (left), Jun\_radiopaedia\_40\_86625\_0\_case18\_53 from the MosMedData+ dataset (middle) and cju2tzypl4wss0799ow05oxb9 from the Kvasir-SEG dataset (right).}
\label{fig:visualization}
\end{figure*}

\subsection{Visualization}

Figure~\ref{fig:visualization} presents a qualitative comparison of ProLearn with state-of-the-art approaches on the QaTa-COV19, MosMedData+ and Kvasir-SEG datasets. As textual availability decreased, existing language-guided segmentation models, which rely heavily on image-text pairs, experienced a pronounced drop in performance. In contrast, ProLearn effectively learned from both image-text and image-only data, demonstrating stable results even under limited textual guidance. The same findings can be observed in saliency maps that pinpoint the regions on which each model was focused. Under decreased text guidance, conventional models produced unstable or diffuse attention patterns, compromising their segmentation accuracy. However, ProLearn, aided by semantic approximation, retained more coherent attention focused on areas of the lesion.

\begin{figure}[t]
  \centering
   \includegraphics[width=0.95\linewidth]{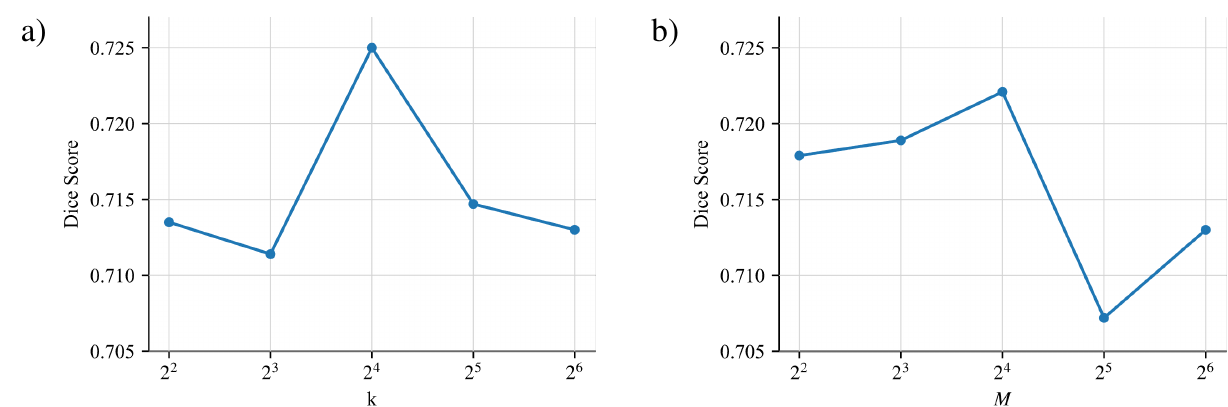}
   \caption{Hyperparameter sensitivity analysis with varying candidate and prototype configurations. a) Effect of varying $k$, number of responding prototype vectors on Dice score; b) Effect of varying $M$, number of prototypes per surrogate label on Dice score.}
\label{fig:hyperparameter}
\end{figure}

\subsection{Hyperparameter Sensitivity}

To investigate the effect of the number of candidates per response $k$ and the number of prototypes per surrogate label $M$, we presented the performance comparison by varying their values for segmentation, as shown in Figure~\ref{fig:hyperparameter}. We adjusted one hyperparameter at a time while keeping the other fixed. The experimental results indicated that our framework remained stable across a broad range of hyperparameter values.

Specifically, $k$ governs the trade-off between information diversity and noise. A small $k$ limits the model's ability to capture subtle segmentation patterns due to insufficient variability. As $k$ increases, richer patterns are incorporated, improving performance. However, when $k$ is too large, irrelevant patterns introduce excessive noise, leading to degraded segmentation accuracy.

For $M$, increasing the number of prototypes enhances performance by capturing diverse segmentation characteristics. However, when $M$ exceeds a certain threshold, prototype redundancy arises, leading to overlapping or irrelevant representations that blur the segmentation boundaries. Thus, balancing $M$ is crucial; too few prototypes hinder representational capacity, while an excessive count dilutes the model's focus on meaningful patterns. 

\section{Conclusion}
\label{sec:conclusion}

We investigated the issue of textual reliance in medical language-guided segmentation, which limits both the applicability of segmentation in clinical workflows and the exploration of image-only training data. To address this issue, we have presented ProLearn, a prototype-driven framework that fundamentally alleviates textual reliance in medical language-guided segmentation. Our experiments demonstrated that ProLearn effectively learns from both image-only and image-text data, making efficient use of the target dataset’s textual descriptions. ProLearn showed stable performance under different text availability and outperformed state-of-the-art image-only approaches under minimal text supervision.


\textbf{Outlook}: PSA can be readily adapted to new medical imaging tasks with few paired reports, enhancing segmentation performance with negligible computational overhead.

{
    \small
    \bibliographystyle{ieeenat_fullname}
    \bibliography{main}

\begin{thebibliography}{49}
\providecommand{\natexlab}[1]{#1}
\providecommand{\url}[1]{\texttt{#1}}
\expandafter\ifx\csname urlstyle\endcsname\relax
  \providecommand{\doi}[1]{doi: #1}\else
  \providecommand{\doi}{doi: \begingroup \urlstyle{rm}\Url}\fi

\bibitem[Azad et~al.(2024)Azad, Aghdam, Rauland, Jia, Avval, Bozorgpour, Karimijafarbigloo, Cohen, Adeli, and Merhof]{unet_review24}
Reza Azad, Ehsan~Khodapanah Aghdam, Amelie Rauland, Yiwei Jia, Atlas~Haddadi Avval, Afshin Bozorgpour, Sanaz Karimijafarbigloo, Joseph~Paul Cohen, Ehsan Adeli, and Dorit Merhof.
\newblock Medical image segmentation review: The success of u-net.
\newblock \emph{IEEE Transactions on Pattern Analysis and Machine Intelligence}, 46\penalty0 (12):\penalty0 10076--10095, 2024.

\bibitem[Butoi et~al.(2023)Butoi, Ortiz, Ma, Sabuncu, Guttag, and Dalca]{universeg}
Victor~Ion Butoi, Jose Javier~Gonzalez Ortiz, Tianyu Ma, Mert~R. Sabuncu, John Guttag, and Adrian~V. Dalca.
\newblock Universeg: Universal medical image segmentation.
\newblock In \emph{Proceedings of the IEEE/CVF International Conference on Computer Vision (ICCV)}, pages 21438--21451, 2023.

\bibitem[Campello et~al.(2013)Campello, Moulavi, and Sander]{hdbscan}
Ricardo~JGB Campello, Davoud Moulavi, and J{\"o}rg Sander.
\newblock Density-based clustering based on hierarchical density estimates.
\newblock In \emph{Pacific-Asia conference on knowledge discovery and data mining}, pages 160--172. Springer, 2013.

\bibitem[Candemir et~al.(2013)Candemir, Jaeger, Palaniappan, Musco, Singh, Xue, Karargyris, Antani, Thoma, and McDonald]{seg_lung}
Sema Candemir, Stefan Jaeger, Kannappan Palaniappan, Jonathan~P Musco, Rahul~K Singh, Zhiyun Xue, Alexandros Karargyris, Sameer Antani, George Thoma, and Clement~J McDonald.
\newblock Lung segmentation in chest radiographs using anatomical atlases with nonrigid registration.
\newblock \emph{IEEE transactions on medical imaging}, 33\penalty0 (2):\penalty0 577--590, 2013.

\bibitem[Cao et~al.(2022)Cao, Wang, Chen, Jiang, Zhang, Tian, and Wang]{swin_unet}
H. Cao, Y. Wang, J. Chen, D. Jiang, X. Zhang, Q. Tian, and M. Wang.
\newblock Swin-unet: Unet-like pure transformer for medical image segmentation.
\newblock In \emph{European conference on computer vision}, pages 205--218. Springer, 2022.

\bibitem[Chen et~al.(2023)Chen, Zhang, Han, Chen, Shi, Xu, and Xu]{vlm_survey23}
Fei-Long Chen, Du-Zhen Zhang, Ming-Lun Han, Xiu-Yi Chen, Jing Shi, Shuang Xu, and Bo Xu.
\newblock Vlp: A survey on vision-language pre-training.
\newblock \emph{Machine Intelligence Research}, 20\penalty0 (1):\penalty0 38--56, 2023.

\bibitem[Chen and Ran(2019)]{edge}
Jiasi Chen and Xukan Ran.
\newblock Deep learning with edge computing: A review.
\newblock \emph{Proceedings of the IEEE}, 107\penalty0 (8):\penalty0 1655--1674, 2019.

\bibitem[Chen et~al.(2021)Chen, Lu, Yu, Luo, Adeli, Wang, Lu, Yuille, and Zhou]{trans_unet}
J. Chen, Y. Lu, Q. Yu, X. Luo, E. Adeli, Y. Wang, L. Lu, A.~L. Yuille, and Y. Zhou.
\newblock Transunet: Transformers make strong encoders for medical image segmentation.
\newblock \emph{arXiv preprint arXiv:2102.04306}, 2021.

\bibitem[Cheng et~al.(2023)Cheng, Lin, Lyu, Huang, Luo, and Tang]{prior}
Pujin Cheng, Li Lin, Junyan Lyu, Yijin Huang, Wenhan Luo, and Xiaoying Tang.
\newblock Prior: Prototype representation joint learning from medical images and reports.
\newblock In \emph{Proceedings of the IEEE/CVF International Conference on Computer Vision}, pages 21361--21371, 2023.

\bibitem[Degerli et~al.(2022)Degerli, Kiranyaz, Chowdhury, and Gabbouj]{qata}
Aysen Degerli, Serkan Kiranyaz, Muhammad E.~H. Chowdhury, and Moncef Gabbouj.
\newblock Osegnet: Operational segmentation network for covid-19 detection using chest x-ray images.
\newblock In \emph{2022 IEEE International Conference on Image Processing (ICIP)}, pages 2306--2310, 2022.

\bibitem[Dice(1945)]{dice}
Lee~R Dice.
\newblock Measures of the amount of ecologic association between species.
\newblock \emph{Ecology}, 26\penalty0 (3):\penalty0 297--302, 1945.

\bibitem[Ferrari et~al.(2012)Ferrari, Carbone, Cappelli, Boni, Melfi, Ferrari, Mosca, and Pietrabissa]{seg_preoperative_planning}
Vincenzo Ferrari, Marina Carbone, Carla Cappelli, Luigi Boni, Franca Melfi, Mauro Ferrari, Franco Mosca, and Andrea Pietrabissa.
\newblock Value of multidetector computed tomography image segmentation for preoperative planning in general surgery.
\newblock \emph{Surgical endoscopy}, 26:\penalty0 616--626, 2012.

\bibitem[Gu et~al.(2022)Gu, Stefani, Wu, Thomason, and Wang]{vlm_sruvey22}
Jing Gu, Eliana Stefani, Qi Wu, Jesse Thomason, and Xin Wang.
\newblock Vision-and-language navigation: A survey of tasks, methods, and future directions.
\newblock In \emph{Proceedings of the 60th Annual Meeting of the Association for Computational Linguistics (Volume 1: Long Papers)}, pages 7606--7623, 2022.

\bibitem[Holden et~al.(2014)Holden, Ungi, Sargent, McGraw, Chen, Ganapathy, Peters, and Fichtinger]{seg_needle}
Matthew~Stephen Holden, Tamas Ungi, Derek Sargent, Robert~C McGraw, Elvis~CS Chen, Sugantha Ganapathy, Terry~M Peters, and Gabor Fichtinger.
\newblock Feasibility of real-time workflow segmentation for tracked needle interventions.
\newblock \emph{IEEE Transactions on Biomedical Engineering}, 61\penalty0 (6):\penalty0 1720--1728, 2014.

\bibitem[Hu et~al.(2024)Hu, Li, Sun, Song, Zhang, Lin, and Chen]{lga}
Jihong Hu, Yinhao Li, Hao Sun, Yu Song, Chujie Zhang, Lanfen Lin, and Yen-Wei Chen.
\newblock Lga: A language guide adapter for advancing the sam model’s capabilities in medical image segmentation.
\newblock page 610–620, Berlin, Heidelberg, 2024. Springer-Verlag.

\bibitem[Huang et~al.(2021)Huang, Shen, Lungren, and Yeung]{gloria}
Shih-Cheng Huang, Liyue Shen, Matthew~P. Lungren, and Serena Yeung.
\newblock Gloria: A multimodal global-local representation learning framework for label-efficient medical image recognition.
\newblock In \emph{2021 IEEE/CVF International Conference on Computer Vision (ICCV)}, pages 3922--3931, 2021.

\bibitem[Jaccard(1901)]{miou}
Paul Jaccard.
\newblock {\'E}tude comparative de la distribution florale dans une portion des alpes et des jura.
\newblock \emph{Bull Soc Vaudoise Sci Nat}, 37:\penalty0 547--579, 1901.

\bibitem[Jha et~al.(2019)Jha, Smedsrud, Riegler, Halvorsen, De~Lange, Johansen, and Johansen]{kvasir}
Debesh Jha, Pia~H Smedsrud, Michael~A Riegler, P{\aa}l Halvorsen, Thomas De~Lange, Dag Johansen, and H{\aa}vard~D Johansen.
\newblock Kvasir-seg: A segmented polyp dataset.
\newblock In \emph{International conference on multimedia modeling}, pages 451--462. Springer, 2019.

\bibitem[Kapur et~al.(2014)Kapur, Egger, Jayender, Toews, and Wells]{seg_therapy}
Tina Kapur, Jan Egger, Jagadeesan Jayender, Matthew Toews, and William~M Wells.
\newblock Registration and segmentation for image-guided therapy.
\newblock \emph{Intraoperative Imaging and Image-Guided Therapy}, pages 79--91, 2014.

\bibitem[Keereweer et~al.(2011)Keereweer, Kerrebijn, Van~Driel, Xie, Kaijzel, Snoeks, Que, Hutteman, Van Der~Vorst, Mieog, et~al.]{seg_sur}
Stijn Keereweer, Jeroen~DF Kerrebijn, Pieter~BAA Van~Driel, Bangwen Xie, Eric~L Kaijzel, Thomas~JA Snoeks, Ivo Que, Merlijn Hutteman, Joost~R Van Der~Vorst, J~Sven~D Mieog, et~al.
\newblock Optical image-guided surgery—where do we stand?
\newblock \emph{Molecular Imaging and Biology}, 13:\penalty0 199--207, 2011.

\bibitem[Li et~al.(2023)Li, Li, Li, Wang, Guo, Lu, Jin, Zhang, and Hong]{lvit}
Zihan Li, Yunxiang Li, Qingde Li, Puyang Wang, Dazhou Guo, Le Lu, Dakai Jin, You Zhang, and Qingqi Hong.
\newblock Lvit: language meets vision transformer in medical image segmentation.
\newblock \emph{IEEE transactions on medical imaging}, 2023.

\bibitem[Lloyd(1982)]{kmeans}
S. Lloyd.
\newblock Least squares quantization in pcm.
\newblock \emph{IEEE Transactions on Information Theory}, 28\penalty0 (2):\penalty0 129--137, 1982.

\bibitem[L{\"u}ddecke and Ecker(2022)]{clip_prompt}
Timo L{\"u}ddecke and Alexander Ecker.
\newblock Image segmentation using text and image prompts.
\newblock In \emph{Proceedings of the IEEE/CVF conference on computer vision and pattern recognition}, pages 7086--7096, 2022.

\bibitem[Ma et~al.(2024)Ma, He, Li, Han, You, and Wang]{medsam}
Jun Ma, Yuting He, Feifei Li, Lin Han, Chenyu You, and Bo Wang.
\newblock Segment anything in medical images.
\newblock \emph{Nature Communications}, 15\penalty0 (1):\penalty0 654, 2024.

\bibitem[Maji et~al.(2022)Maji, Sigedar, and Singh]{seg_brain}
Dhiraj Maji, Prarthana Sigedar, and Munendra Singh.
\newblock Attention res-unet with guided decoder for semantic segmentation of brain tumors.
\newblock \emph{Biomedical Signal Processing and Control}, 71:\penalty0 103077, 2022.

\bibitem[Michael et~al.(2021)Michael, Ma, Li, Kulwa, and Li]{seg_breast}
Epimack Michael, He Ma, Hong Li, Frank Kulwa, and Jing Li.
\newblock Breast cancer segmentation methods: current status and future potentials.
\newblock \emph{BioMed research international}, 2021\penalty0 (1):\penalty0 9962109, 2021.

\bibitem[Morozov et~al.(2020)Morozov, Andreychenko, Pavlov, Vladzymyrskyy, Ledikhova, Gombolevskiy, Blokhin, Gelezhe, Gonchar, Chernina, et~al.]{mosmeddata}
Sergey Morozov, Anna Andreychenko, Nikolay Pavlov, Anton Vladzymyrskyy, Natalya Ledikhova, Victor Gombolevskiy, Ivan Blokhin, Pavel Gelezhe, Anna Gonchar, Valeria Chernina, et~al.
\newblock Mosmeddata: Chest ct scans with covid-19 related findings.
\newblock 2020.

\bibitem[Oktay et~al.(2018)Oktay, Schlemper, Folgoc, Lee, Heinrich, Misawa, Mori, McDonagh, Hammerla, Kainz, and et~al.]{att_unet}
O. Oktay, J. Schlemper, L.~L. Folgoc, M. Lee, M. Heinrich, K. Misawa, K. Mori, S. McDonagh, N.~Y. Hammerla, B. Kainz, and et al.
\newblock Attention u-net: Learning where to look for the pancreas.
\newblock \emph{arXiv preprint arXiv:1804.03999}, 2018.

\bibitem[Qiu et~al.(2024)Qiu, Wu, Zhang, Lin, Wang, Zhang, Wang, and Xie]{llama3-med}
Pengcheng Qiu, Chaoyi Wu, Xiaoman Zhang, Weixiong Lin, Haicheng Wang, Ya Zhang, Yanfeng Wang, and Weidi Xie.
\newblock Towards building multilingual language model for medicine.
\newblock \emph{Nature Communications}, 15\penalty0 (1):\penalty0 8384, 2024.

\bibitem[Radford et~al.(2019)Radford, Wu, Child, Luan, Amodei, and Sutskever]{gpt2}
Alec Radford, Jeff Wu, Rewon Child, David Luan, Dario Amodei, and Ilya Sutskever.
\newblock Language models are unsupervised multitask learners.
\newblock 2019.

\bibitem[Radford et~al.(2021)Radford, Kim, Hallacy, Ramesh, Goh, Agarwal, Sastry, Askell, Mishkin, Clark, et~al.]{clip}
Alec Radford, Jong~Wook Kim, Chris Hallacy, Aditya Ramesh, Gabriel Goh, Sandhini Agarwal, Girish Sastry, Amanda Askell, Pamela Mishkin, Jack Clark, et~al.
\newblock Learning transferable visual models from natural language supervision.
\newblock In \emph{International conference on machine learning}, pages 8748--8763. PMLR, 2021.

\bibitem[Ronneberger et~al.(2015)Ronneberger, Fischer, and Brox]{unet}
O. Ronneberger, P. Fischer, and T. Brox.
\newblock U-net: Convolutional networks for biomedical image segmentation.
\newblock In \emph{Medical Image Computing and Computer-Assisted Intervention--MICCAI 2015: 18th International Conference, Munich, Germany, October 5-9, 2015, Proceedings, Part III 18}, pages 234--241. Springer, 2015.

\bibitem[Siddique et~al.(2021)Siddique, Paheding, Elkin, and Devabhaktuni]{unet_review21}
N. Siddique, S. Paheding, C.~P. Elkin, and V. Devabhaktuni.
\newblock U-net and its variants for medical image segmentation: A review of theory and applications.
\newblock \emph{IEEE Access}, 9:\penalty0 82031--82057, 2021.

\bibitem[Snell et~al.(2017)Snell, Swersky, and Zemel]{prototype_fewshot}
Jake Snell, Kevin Swersky, and Richard Zemel.
\newblock Prototypical networks for few-shot learning.
\newblock \emph{Advances in neural information processing systems}, 30, 2017.

\bibitem[Thirunavukarasu et~al.(2023)Thirunavukarasu, Ting, Elangovan, Gutierrez, Tan, and Ting]{llm_med}
Arun~James Thirunavukarasu, Darren Shu~Jeng Ting, Kabilan Elangovan, Laura Gutierrez, Ting~Fang Tan, and Daniel Shu~Wei Ting.
\newblock Large language models in medicine.
\newblock \emph{Nature medicine}, 29\penalty0 (8):\penalty0 1930--1940, 2023.

\bibitem[Van Den~Oord et~al.(2017)Van Den~Oord, Vinyals, et~al.]{discrete_rl}
Aaron Van Den~Oord, Oriol Vinyals, et~al.
\newblock Neural discrete representation learning.
\newblock \emph{Advances in neural information processing systems}, 30, 2017.

\bibitem[Wang et~al.(2022)Wang, Lu, Li, Tao, Guo, Gong, and Liu]{clip_ref}
Zhaoqing Wang, Yu Lu, Qiang Li, Xunqiang Tao, Yandong Guo, Mingming Gong, and Tongliang Liu.
\newblock Cris: Clip-driven referring image segmentation.
\newblock In \emph{Proceedings of the IEEE/CVF conference on computer vision and pattern recognition}, pages 11686--11695, 2022.

\bibitem[Wijewickrema et~al.(2016)Wijewickrema, Zhou, Bailey, Kennedy, and O'Leary]{seg_vr}
Sudanthi Wijewickrema, Yun Zhou, James Bailey, Gregor Kennedy, and Stephen O'Leary.
\newblock Provision of automated step-by-step procedural guidance in virtual reality surgery simulation.
\newblock In \emph{Proceedings of the 22nd ACM Conference on Virtual Reality Software and Technology}, pages 69--72, 2016.

\bibitem[Xu et~al.(2023)Xu, Chen, Zhang, Song, Wan, and Li]{clip_seg_tune}
Zunnan Xu, Zhihong Chen, Yong Zhang, Yibing Song, Xiang Wan, and Guanbin Li.
\newblock Bridging vision and language encoders: Parameter-efficient tuning for referring image segmentation.
\newblock In \emph{Proceedings of the IEEE/CVF International Conference on Computer Vision}, pages 17503--17512, 2023.

\bibitem[Yang et~al.(2018)Yang, Zhang, Yin, and Liu]{prototype_class}
Hong-Ming Yang, Xu-Yao Zhang, Fei Yin, and Cheng-Lin Liu.
\newblock Robust classification with convolutional prototype learning.
\newblock In \emph{Proceedings of the IEEE conference on computer vision and pattern recognition}, pages 3474--3482, 2018.

\bibitem[Yang et~al.(2020)Yang, Liu, Liu, Zhang, Wan, Huang, Chen, and Zhang]{covid}
Li Yang, Shasha Liu, Jinyan Liu, Zhixin Zhang, Xiaochun Wan, Bo Huang, Youhai Chen, and Yi Zhang.
\newblock Covid-19: immunopathogenesis and immunotherapeutics.
\newblock \emph{Signal transduction and targeted therapy}, 5\penalty0 (1):\penalty0 128, 2020.

\bibitem[Ye et~al.(2024)Ye, Meng, Li, Feng, and Kim]{sgseg}
Shuchang Ye, Mingyuan Meng, Mingjian Li, Dagan Feng, and Jinman Kim.
\newblock Enabling text-free inference in language-guided segmentation of chest x-rays via self-guidance.
\newblock In \emph{International Conference on Medical Image Computing and Computer-Assisted Intervention}, pages 242--252. Springer, 2024.

\bibitem[Yin et~al.(2022)Yin, Sun, Fu, Lu, and Zhang]{unet_review22}
Xiao-Xia Yin, Le Sun, Yuhan Fu, Ruiliang Lu, and Yanchun Zhang.
\newblock [retracted] u-net-based medical image segmentation.
\newblock \emph{Journal of healthcare engineering}, 2022\penalty0 (1):\penalty0 4189781, 2022.

\bibitem[Zhang et~al.(2024{\natexlab{a}})Zhang, Huang, Jin, and Lu]{vlm_survey24}
Jingyi Zhang, Jiaxing Huang, Sheng Jin, and Shijian Lu.
\newblock Vision-language models for vision tasks: A survey.
\newblock \emph{IEEE Transactions on Pattern Analysis and Machine Intelligence}, 2024{\natexlab{a}}.

\bibitem[Zhang et~al.(2025)Zhang, Xu, Usuyama, Xu, Bagga, Tinn, Preston, Rao, Wei, Valluri, Wong, Tupini, Wang, Mazzola, Shukla, Liden, Gao, Crabtree, Piening, Bifulco, Lungren, Naumann, Wang, and Poon]{biomedclip}
Sheng Zhang, Yanbo Xu, Naoto Usuyama, Hanwen Xu, Jaspreet Bagga, Robert Tinn, Sam Preston, Rajesh Rao, Mu Wei, Naveen Valluri, Cliff Wong, Andrea Tupini, Yu Wang, Matt Mazzola, Swadheen Shukla, Lars Liden, Jianfeng Gao, Angela Crabtree, Brian Piening, Carlo Bifulco, Matthew~P. Lungren, Tristan Naumann, Sheng Wang, and Hoifung Poon.
\newblock A multimodal biomedical foundation model trained from fifteen million image–text pairs.
\newblock \emph{NEJM AI}, 2\penalty0 (1):\penalty0 AIoa2400640, 2025.

\bibitem[Zhang et~al.(2024{\natexlab{b}})Zhang, Ni, Yang, and Zhang]{madapter}
Xu Zhang, Bo Ni, Yang Yang, and Lefei Zhang.
\newblock Madapter: A better interaction between image and language for medical image segmentation.
\newblock In \emph{International Conference on Medical Image Computing and Computer-Assisted Intervention}, pages 425--434. Springer, 2024{\natexlab{b}}.

\bibitem[Zhao et~al.(2025)Zhao, Gu, Yang, Usuyama, Lee, Kiblawi, Naumann, Gao, Crabtree, Abel, Moung-Wen, Piening, Bifulco, Wei, Poon, and Wang]{biomedparse}
Theodore Zhao, Yu Gu, Jianwei Yang, Naoto Usuyama, Hin Lee, Ho\, Sid Kiblawi, Tristan Naumann, Jianfeng Gao, Angela Crabtree, Jacob Abel, Christine Moung-Wen, Brian Piening, Carlo Bifulco, Mu Wei, Hoifung Poon, and Sheng Wang.
\newblock A foundation model for joint segmentation, detection and recognition of biomedical objects across nine modalities.
\newblock \emph{Nature Methods}, 22\penalty0 (1):\penalty0 166--176, 2025.

\bibitem[Zhong et~al.(2023)Zhong, Xu, Liang, Chen, and Wu]{guidedecoder}
Yi Zhong, Mengqiu Xu, Kongming Liang, Kaixin Chen, and Ming Wu.
\newblock Ariadne’s thread: Using text prompts to improve segmentation of infected areas from chest x-ray images.
\newblock In \emph{International Conference on Medical Image Computing and Computer-Assisted Intervention}, pages 724--733. Springer, 2023.

\bibitem[Zhou et~al.(2018)Zhou, Siddiquee, Tajbakhsh, and Liang]{unet++}
Z. Zhou, M.~M.~Rahman Siddiquee, N. Tajbakhsh, and J. Liang.
\newblock Unet++: A nested u-net architecture for medical image segmentation.
\newblock In \emph{Deep Learning in Medical Image Analysis and Multimodal Learning for Clinical Decision Support: 4th International Workshop, DLMIA 2018, and 8th International Workshop, ML-CDS 2018, Held in Conjunction with MICCAI 2018, Granada, Spain, September 20, 2018, Proceedings 4}, pages 3--11. Springer, 2018.

\end{thebibliography}
}

\clearpage

\appendix

\newcommand{\appendixfigures}{
    \renewcommand{\thefigure}{A\arabic{figure}}
    \setcounter{figure}{\value{appendixfig}}
}
\newcounter{appendixfig}

\newcommand{\appendixtables}{
    \renewcommand{\thetable}{A\arabic{table}}
    \setcounter{table}{\value{appendixtab}}
}
\newcounter{appendixtab}

\newcommand{\appendixequations}{
    \renewcommand{\theequation}{A\arabic{equation}}
    \setcounter{equation}{\value{appendixequation}}
}
\newcounter{appendixequation}

\appendixfigures
\appendixtables

\section{Appendix / Supplemental Material}

\subsection{Pseudocode}
\label{ap:pseudocode}

In this section, we provide pseudocodes to illustrate the workflow of PSA. Algorithm~\ref{alg:PSA_initialization} presents the initialization process, detailing how discrete prototypes are identified from paired image–text data, while Algorithm~\ref{alg:PSA_query_response} demonstrates the query–response mechanism generating approximate textual semantics purely from image embeddings.

\begin{algorithm}[H]
\caption{PSA Initialization}
\label{alg:PSA_initialization}
\begin{algorithmic}[1]
\STATE \textbf{Define}: 
\STATE \quad - A set of $K$ paired samples $\{ (I_i, T_i) \}_{i=1}^K$; 
\STATE \quad - image-only samples $\{ I_j \}$;
\STATE \quad - pretrained encoders $f_{\text{enc}}^I$ and $f_{\text{enc}}^T$;
\STATE \quad - cross-attention module (Language-Guided U-Net)
\STATE \quad - token selection threshold $\tau$;
\STATE \quad - number of semantic clusters $N$ (HDBSCAN);
\STATE \quad - number of sub-clusters $M$ (K-means).
\STATE Return: Prototype space $\mathcal{S} = (\mathcal{S}^Q, \mathcal{S}^R)$

\STATE \textbf{Step 1: Encode Paired Samples}
\FOR {$i = 1$ to $K$}
  \STATE $e^I_i \leftarrow f_{\text{enc}}^I(I_i)$
  \STATE $e^T_i \leftarrow f_{\text{enc}}^T(T_i)$
\ENDFOR

\STATE \textbf{Step 2: Extract Segmentation-Relevant Tokens}
\FOR {$i = 1$ to $K$}
  \STATE Compute cross-attention scores $\alpha_j$ for each token $t_j$ in $T_i$
  \STATE $T_i^{\text{selected}} \leftarrow \{ t_j \mid \alpha_j > \tau\}$
  \STATE $e_i^{\text{sem}} \leftarrow f_{\text{enc}}^T(T_i^{\text{selected}})$
\ENDFOR

\STATE \textbf{Step 3: Cluster Textual Semantics (HDBSCAN)}
\STATE Perform HDBSCAN on $\{ e_i^{\text{sem}} \}$ to form $N$ clusters $\{\mathcal{C}_1, \dots, \mathcal{C}_N\}$

\STATE \textbf{Step 4: Form Image Sub-Clusters (K-means)}
\STATE $\mathcal{S}^Q \leftarrow \emptyset, \quad \mathcal{S}^R \leftarrow \emptyset$
\FOR {$i = 1$ to $N$}
  \STATE Extract embeddings $\{ (e^I_j, e^T_j) \mid j \in \mathcal{C}_i\}$
  \STATE Run K-means with $M$ sub-clusters: $\mathcal{C}_{i1}, \dots, \mathcal{C}_{iM}$
  \FOR {$j = 1$ to $M$}
    \STATE Identify representative $c_{ij} = (e^I_k, e^T_k)$ closest to sub-cluster centroid
    \STATE $q_{ij} \leftarrow e^I_k$ \quad (query prototype)
    \STATE $r_{ij} \leftarrow e^T_k$ \quad (response prototype)
    \STATE $\mathcal{S}^Q \leftarrow \mathcal{S}^Q \cup \{ q_{ij} \}, \quad \mathcal{S}^R \leftarrow \mathcal{S}^R \cup \{ r_{ij} \}$
  \ENDFOR
\ENDFOR

\STATE \textbf{Step 5: Output Prototype Space}
\STATE $\mathcal{S} \leftarrow (\mathcal{S}^Q, \mathcal{S}^R)$
\RETURN $\mathcal{S}$
\end{algorithmic}
\end{algorithm}

\begin{algorithm}[t]
\caption{PSA Query and Response}
\label{alg:PSA_query_response}
\begin{algorithmic}[1]
\REQUIRE 
  Prototype space $\mathcal{S} = (\mathcal{S}^Q, \mathcal{S}^R)$;
  pretrained image encoder $f_{\text{enc}}^I$;
  Language-Guided U-Net $f_{\text{seg}}$;
  query image $I^*$;
  top-$k$ integer $k$.
\ENSURE Approximated textual feature $r^*$ for guiding segmentation

\STATE \textbf{Step 1: Encode the Query Image}
\STATE $q^* \leftarrow f_{\text{enc}}^I(I^*)$

\STATE \textbf{Step 2: Compute Similarity Scores}
\FORALL {$q_{ij}$ in $\mathcal{S}^Q$}
  \STATE $s_{ij} \leftarrow \operatorname{cosine\_similarity}(q^*, q_{ij})$
\ENDFOR

\STATE \textbf{Step 3: Select Top-$k$ Queries}
\STATE $Q^* \leftarrow \operatorname{arg\,top}_k(\{s_{ij}\})$

\STATE \textbf{Step 4: Retrieve Corresponding Responses}
\STATE $R^* \leftarrow \{ r_{ij} \mid q_{ij} \in Q^*\}$

\STATE \textbf{Step 5: Aggregate Responses (Weighted Sum)}
\STATE $r^* \leftarrow \sum_{(q_{ij},\,r_{ij}) \in Q^* \times R^*} w_{ij} \, r_{ij}$ 
\STATE \textbf{where} $w_{ij} = \frac{\exp(s_{ij})}{\sum_{q_{i'j'} \in Q^*} \exp(s_{i'j'})}$

\RETURN $r^*$
\end{algorithmic}
\end{algorithm}

\subsection{Implementation Details}

Following the previous design of language-guided segmentation networks~\cite{guidedecoder, sgseg}, we adopt a U-Net backbone with feature fusion at the decoder stage. The image is resized into $224 \times 224$, and textual reports are tokenized, truncated, and padded to a fixed length of $256$ tokens. To construct the prototype space we set the number of surrogate labels to $6$, with each label containing $64$ prototypes. During inference, the PSA module retrieves the top $10$ prototype candidates per query for semantic approximation. We use the AdamW optimizer with an initial learning rate of $10^{-4}$, which is scheduled to decay using cosine annealing.

\subsection{Limitations and Future Works}

In this work, we focused on demonstrating the core idea of ProLearn in single-label 2D segmentation. Future directions involve exploring multi-label and volumetric data, broader imaging modalities, and extending to more language-guided vision tasks.

\subsection{Visualization}
\label{ap:visualization}

To further demonstrate the effectiveness of our proposed ProLearn, we provide additional visual comparisons of segmentation results in the next page. Specifically, we show the performance of LViT, GuideSeg, SGSeg, and our ProLearn on the QaTa-COV19 and MosMedData+ dataset under different text availability ($1\%$, $5\%$, $10\%$, $25\%$, and $50\%$).

\clearpage

\begin{figure}[H]
  \centering
   \includegraphics[width=0.95\linewidth]{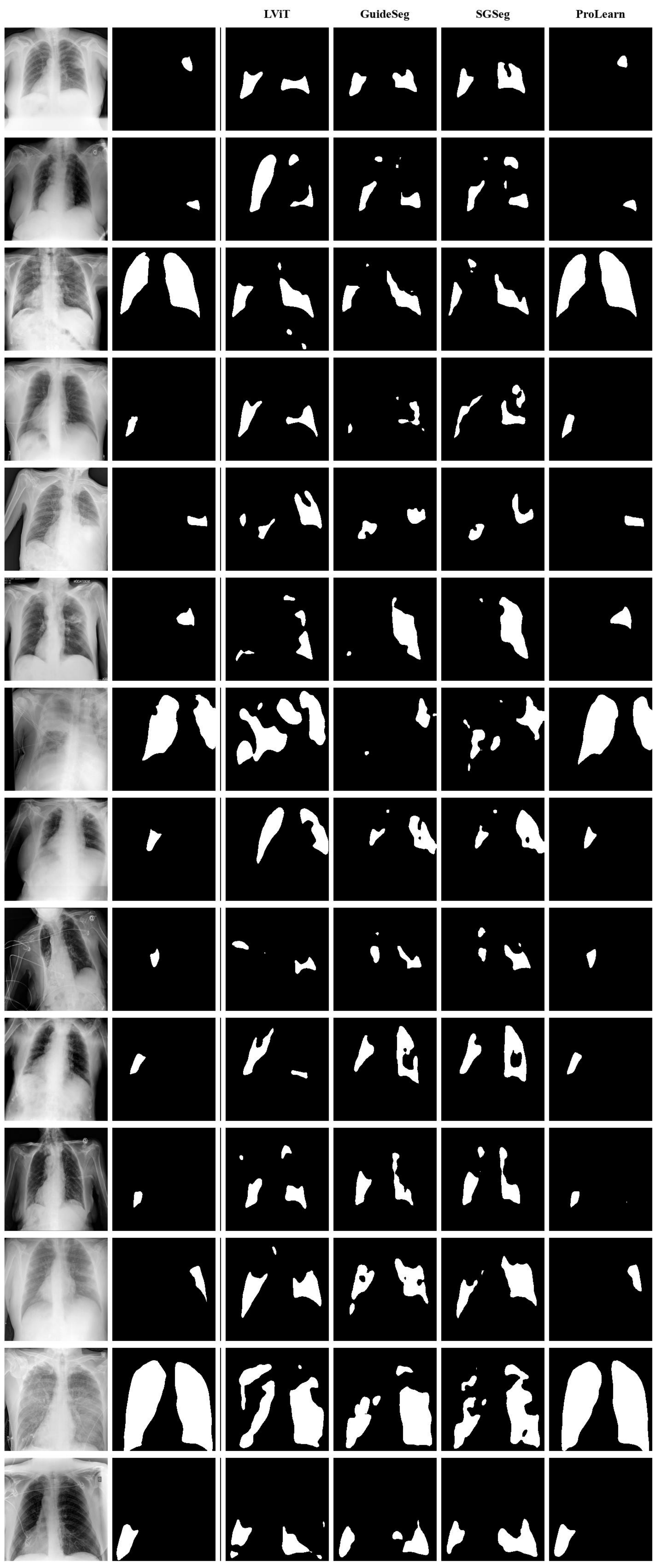}
   \caption{Comparison of segmentation results among LViT, GuideSeg, SGSeg, and our ProLearn on QaTa-COV19 under $1\%$ text availability.}
\label{fig:vis_qata_0.01}
\end{figure}

\begin{figure}[H]
  \centering
   \includegraphics[width=0.95\linewidth]{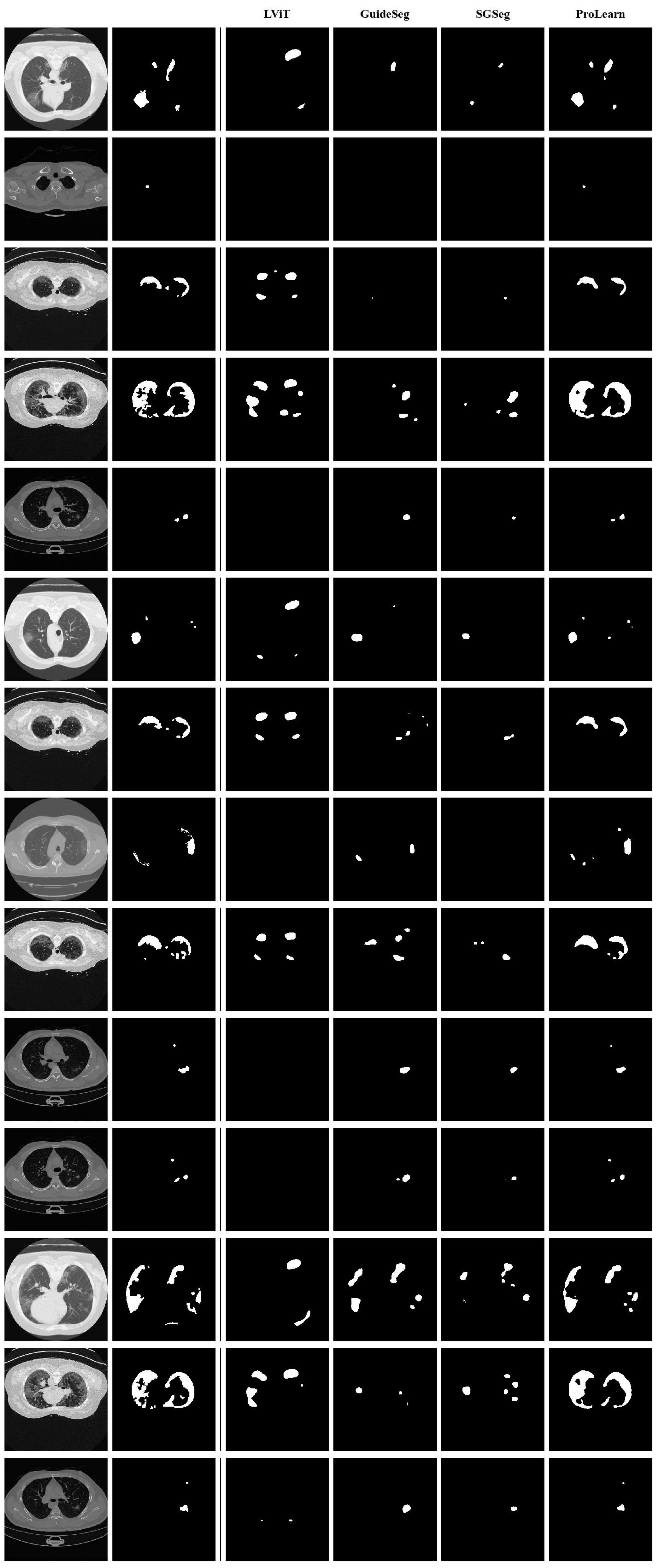}
   \caption{Comparison of segmentation results among LViT, GuideSeg, SGSeg, and our ProLearn on MosMedData+ dataset under $1\%$ text availability.}
\label{fig:vis_mosmeddata_0.01}
\end{figure}

\begin{figure}[H]
  \centering
   \includegraphics[width=0.95\linewidth]{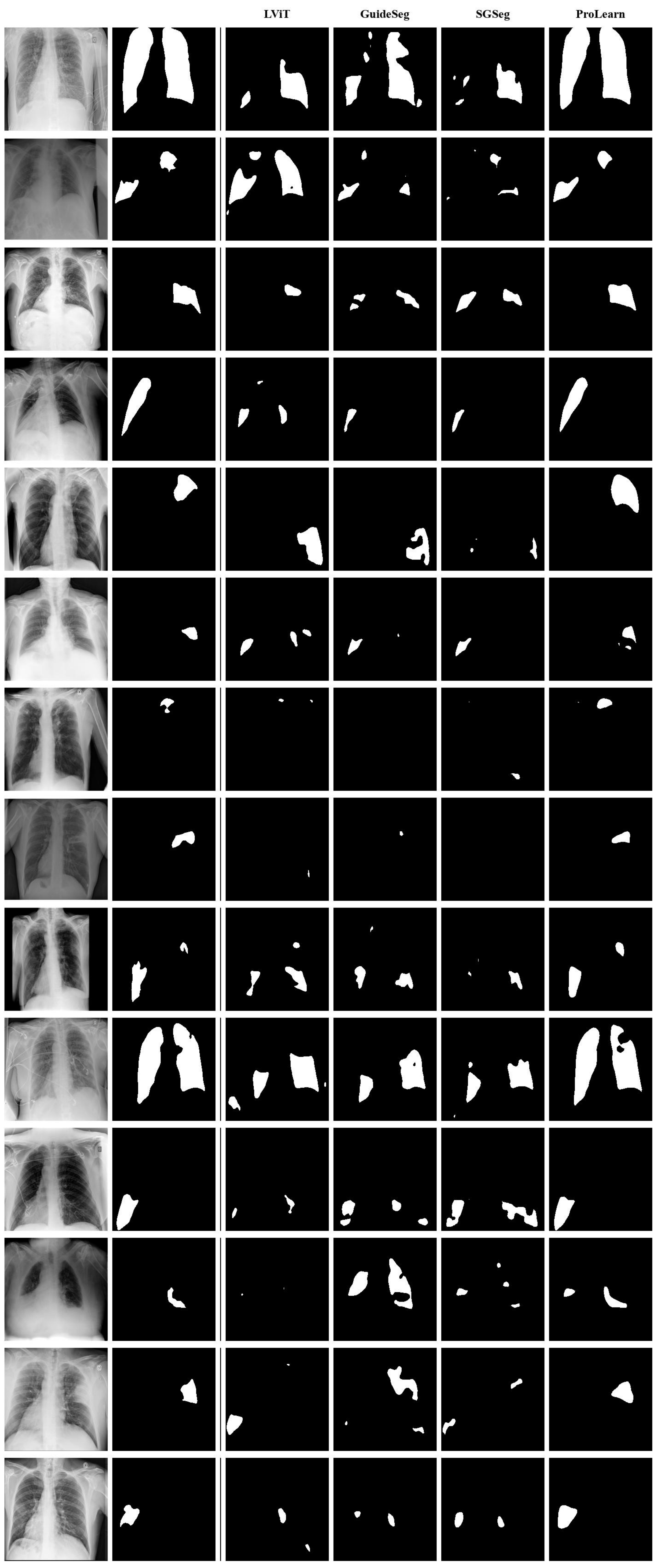}
   \caption{Comparison of segmentation results among LViT, GuideSeg, SGSeg, and our ProLearn on QaTa-COV19 dataset under $5\%$ text availability.}
\label{fig:vis_qata_0.05}
\end{figure}

\begin{figure}[H]
  \centering
   \includegraphics[width=0.95\linewidth]{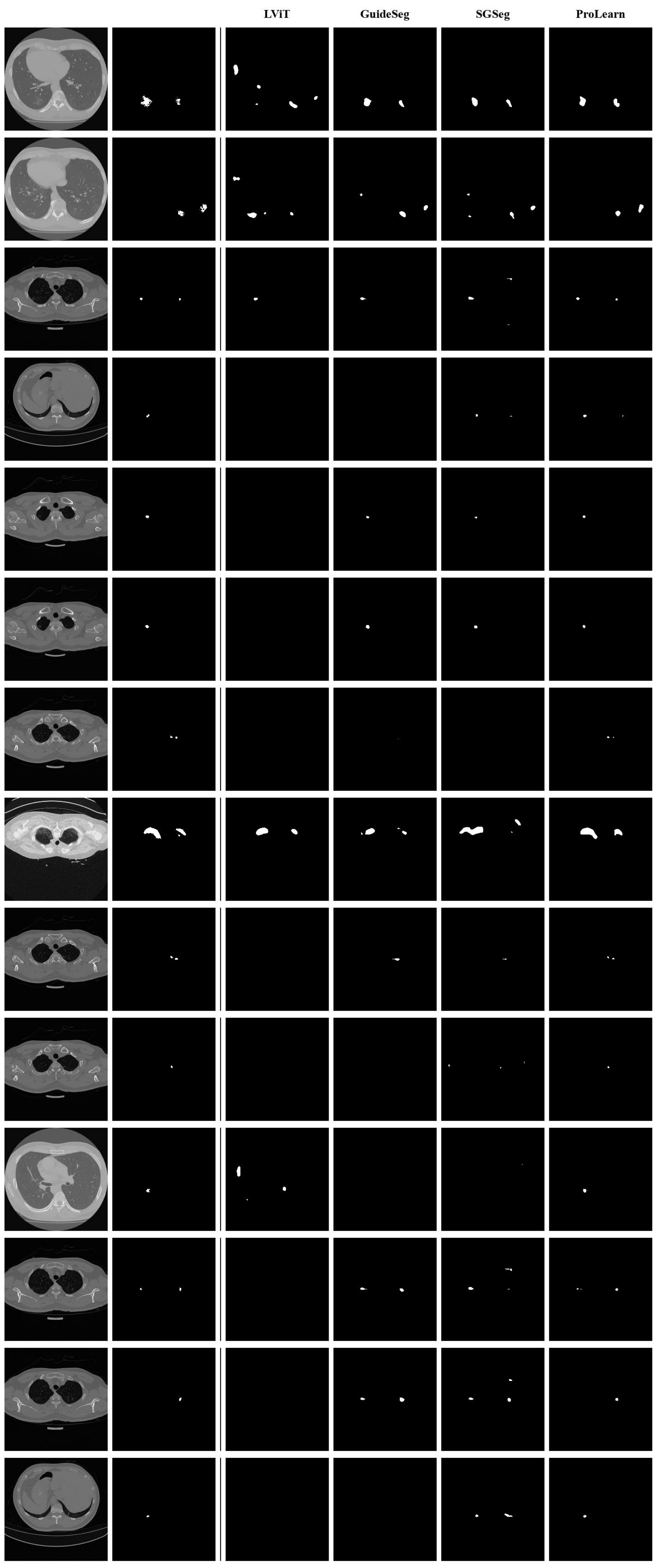}
   \caption{Comparison of segmentation results among LViT, GuideSeg, SGSeg, and our ProLearn on MosMedData+ dataset under $5\%$ text availability.}
\label{fig:vis_mosmeddata_0.05}
\end{figure}

\begin{figure}[H]
  \centering
   \includegraphics[width=0.95\linewidth]{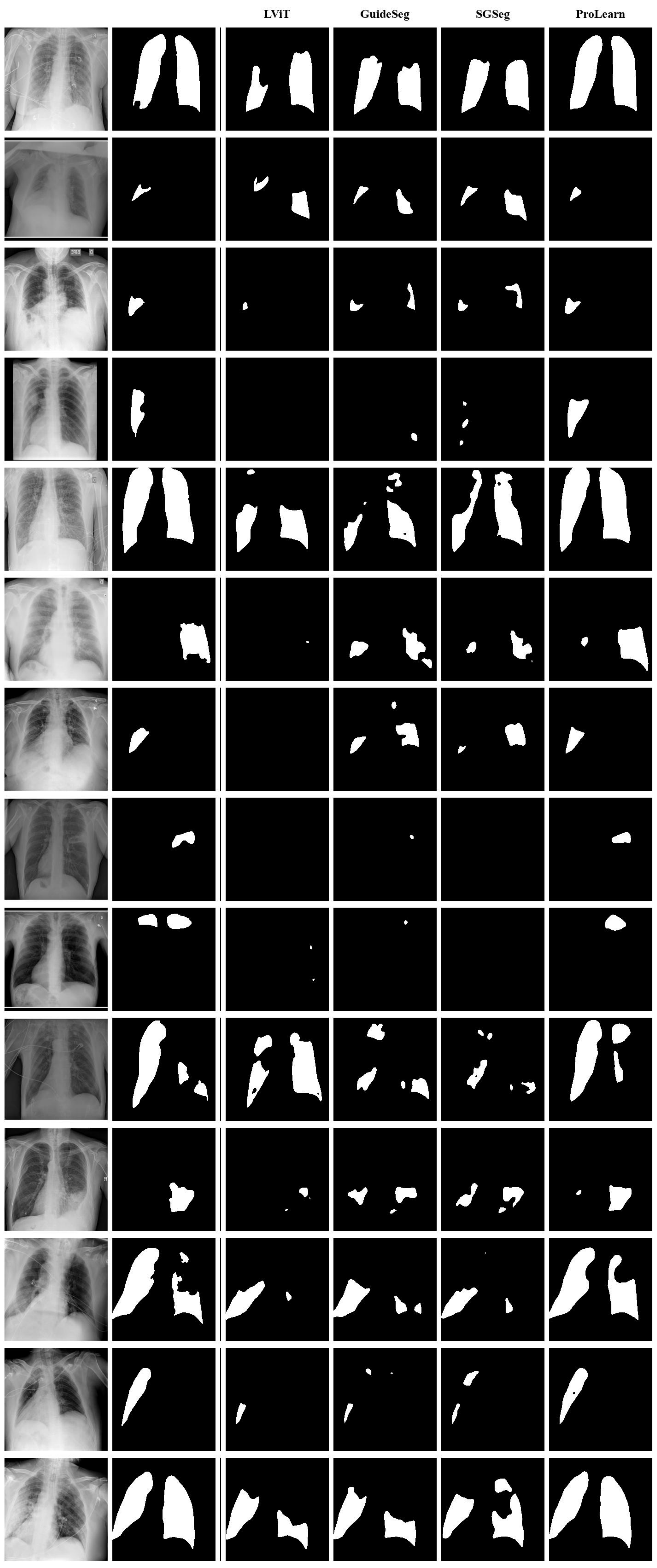}
   \caption{Comparison of segmentation results among LViT, GuideSeg, SGSeg, and our ProLearn on QaTa-COV19 dataset under $10\%$ text availability.}
\label{fig:vis_qata_0.1}
\end{figure}

\begin{figure}[H]
  \centering
   \includegraphics[width=0.95\linewidth]{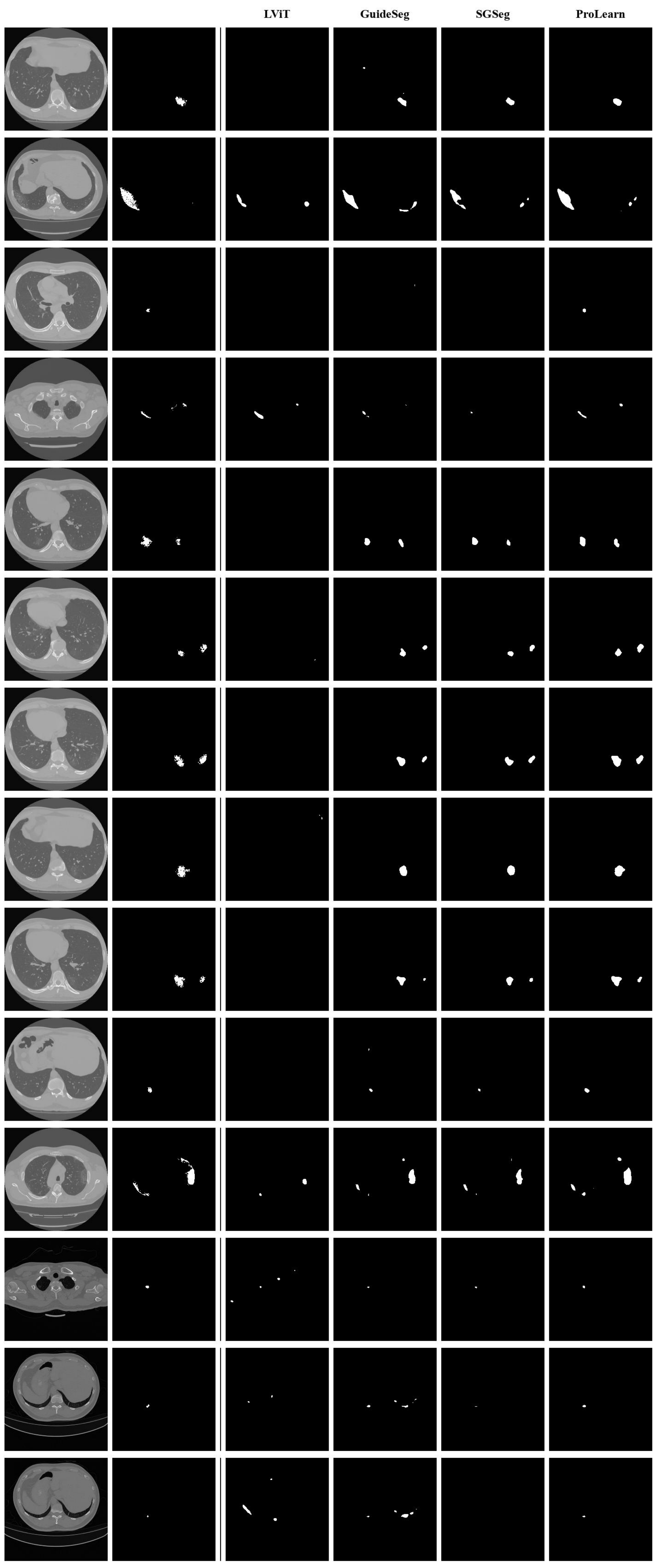}
   \caption{Comparison of segmentation results among LViT, GuideSeg, SGSeg, and our ProLearn on MosMedData+ dataset under $10\%$ text availability.}
\label{fig:vis_mosmeddata_0.1}
\end{figure}

\begin{figure}[H]
  \centering
   \includegraphics[width=0.95\linewidth]{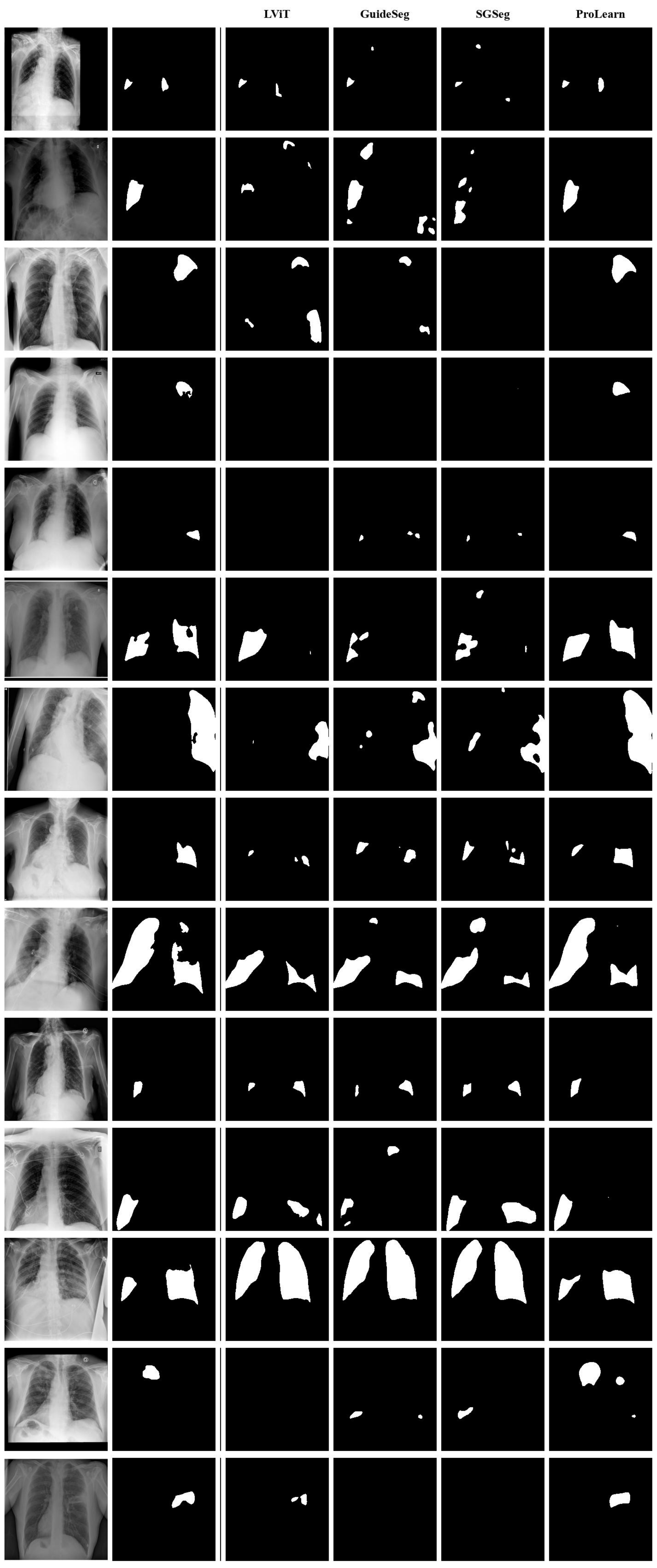}
   \caption{Comparison of segmentation results among LViT, GuideSeg, SGSeg, and our ProLearn on QaTa-COV19 dataset under $25\%$ text availability.}
\label{fig:vis_qata_0.25}
\end{figure}

\begin{figure}[H]
  \centering
   \includegraphics[width=0.95\linewidth]{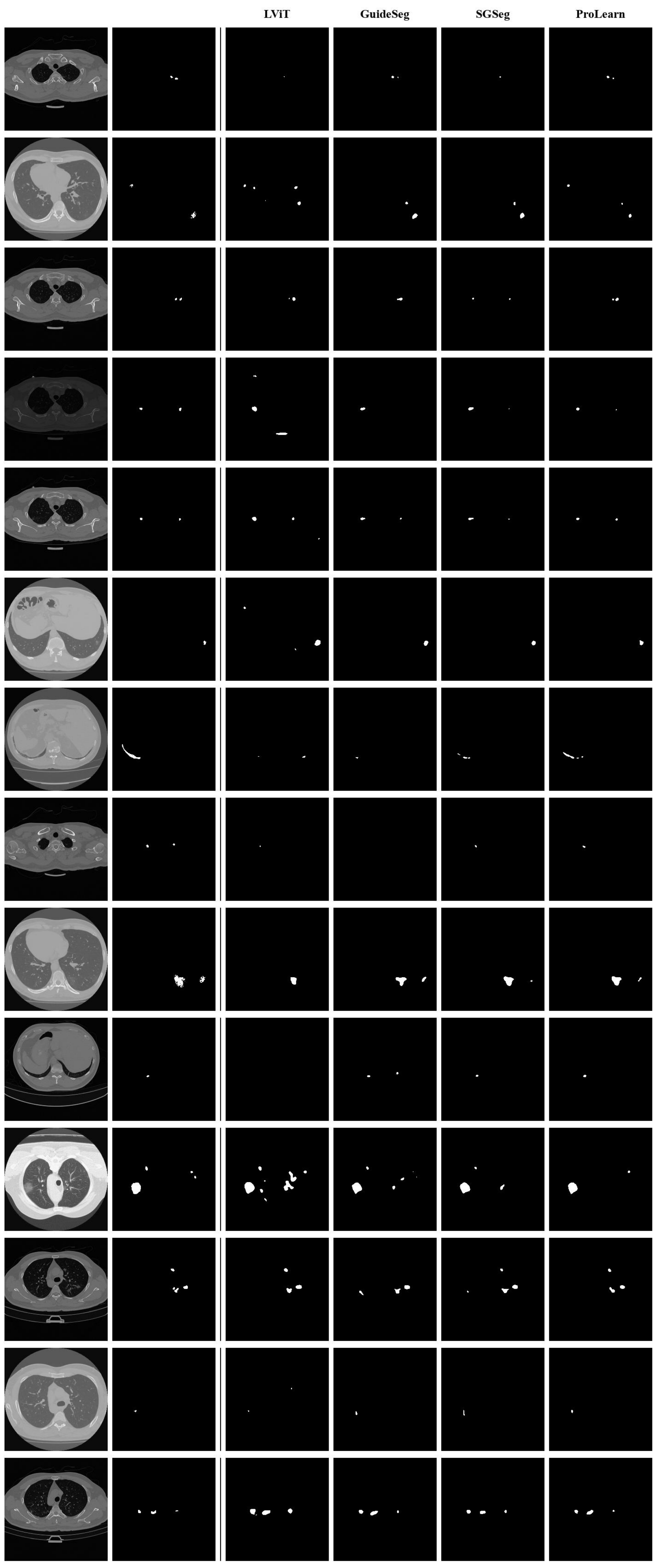}
   \caption{Comparison of segmentation results among LViT, GuideSeg, SGSeg, and our ProLearn on MosMedData+ dataset under $25\%$ text availability.}
\label{fig:vis_mosmeddata_0.25}
\end{figure}

\begin{figure}[H]
  \centering
   \includegraphics[width=0.95\linewidth]{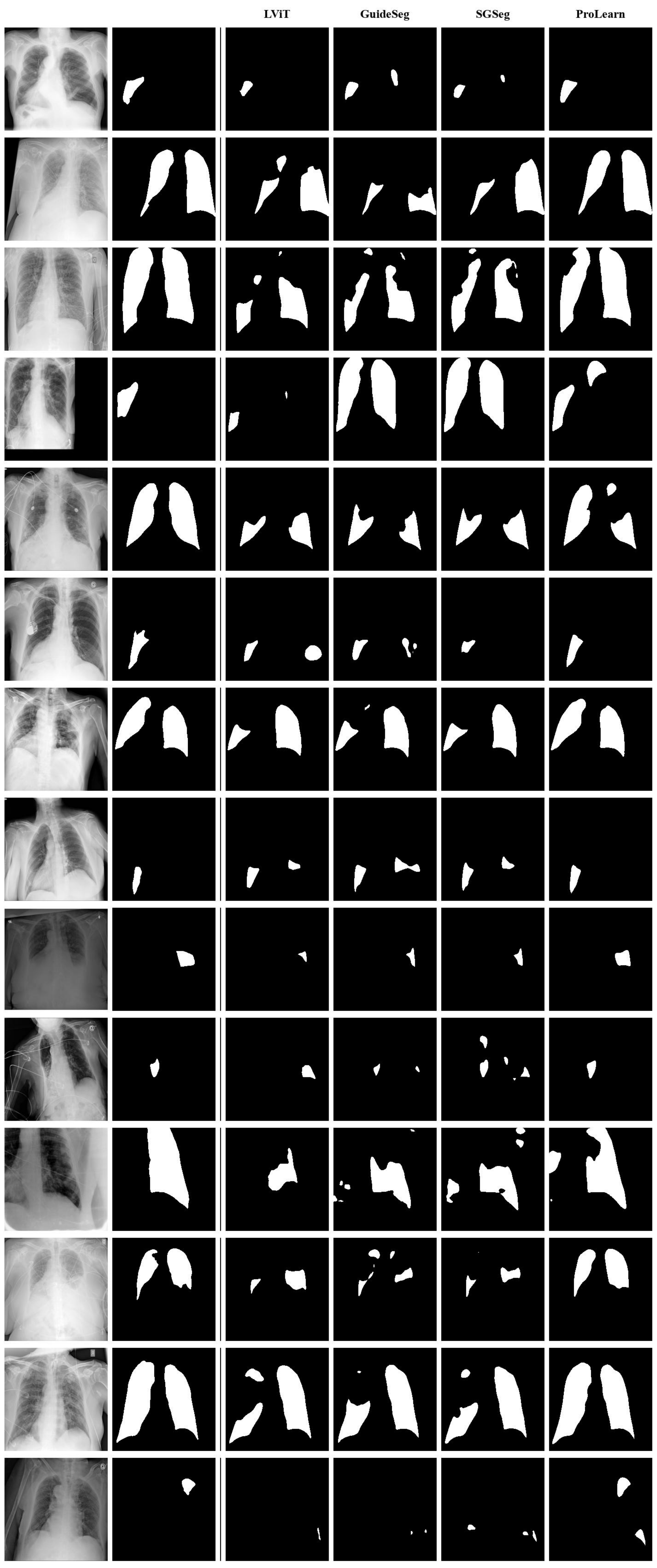}
   \caption{Comparison of segmentation results among LViT, GuideSeg, SGSeg, and our ProLearn on QaTa-COV19 dataset under $5\%$ text availability.}
\label{fig:vis_qata_0.5}
\end{figure}

\begin{figure}[H]
  \centering
   \includegraphics[width=0.95\linewidth]{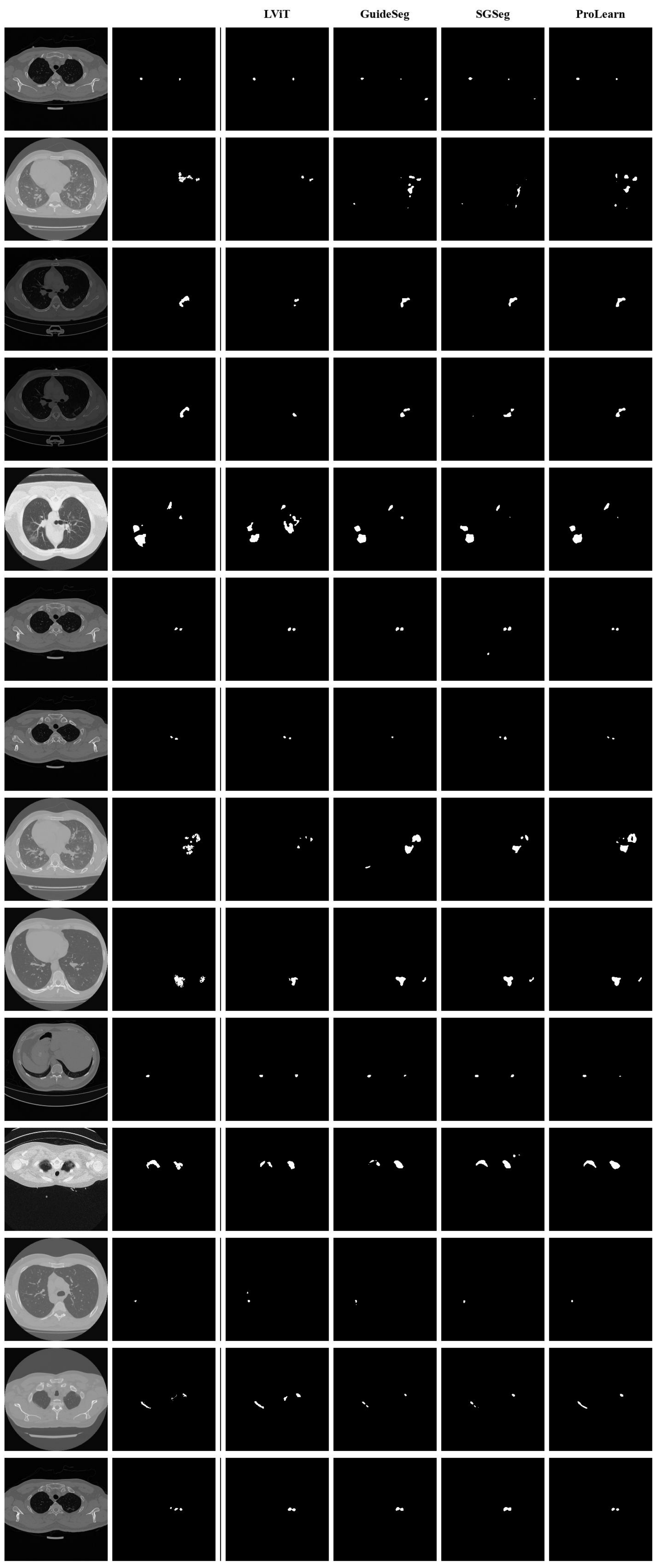}
   \caption{Comparison of segmentation results among LViT, GuideSeg, SGSeg, and our ProLearn on MosMedData+ dataset under $5\%$ text availability.}
\label{fig:vis_mosmeddata_0.5}
\end{figure}

\end{document}